\documentclass[oumk,xs]{xxllp}
\usepackage{amsmath,amssymb, amsfonts} 

\usepackage[TS1,T1]{fontenc}
\usepackage{xxllp}
\usepackage{LLPthm}
\usepackage[mathb]{xlmodern}  
\usepackage{mathrsfs}

\usepackage[breaklinks,colorlinks,citecolor=blue,urlcolor=blue]{hyperref}

\usepackage[natbibapa]{apacite}
\renewcommand{\pubonline}{XXXXX XX, 202X}
\newcommand*{\mies}{XXXXXX}
\renewcommand\figurename{\small Figure}
\renewcommand\tablename{\small Table}
\newtheorem{theorem}{Theorem}
\newtheorem{lemma}{Lemma}
\newtheorem{fact}{Fact}
\newtheorem{corollary}{Corollary}

\theoremstyle{remark}

\newtheorem{example}{Example}

\theoremstyle{definition}
\newtheorem{definition}{Definition}

\renewcommand{\qedsymbol}{$\blacksquare$}
\usepackage{turnstile}
\usepackage{ebproof}
\usepackage{float}
\usepackage[hang,flushmargin]{footmisc} 
\floatstyle{boxed} 
\restylefloat{figure}
\usepackage{pdflscape}
\DeclareFontFamily{U} {MnSymbolA}{}

\DeclareFontShape{U}{MnSymbolA}{m}{n}{
	<-6> MnSymbolA5
	<6-7> MnSymbolA6
	<7-8> MnSymbolA7
	<8-9> MnSymbolA8
	<9-10> MnSymbolA9
	<10-12> MnSymbolA10
	<12-> MnSymbolA12}{}
\DeclareFontShape{U}{MnSymbolA}{b}{n}{
	<-6> MnSymbolA-Bold5
	<6-7> MnSymbolA-Bold6
	<7-8> MnSymbolA-Bold7
	<8-9> MnSymbolA-Bold8
	<9-10> MnSymbolA-Bold9
	<10-12> MnSymbolA-Bold10
	<12-> MnSymbolA-Bold12}{}

\DeclareSymbolFont{MnSyA}{U}{MnSymbolA}{m}{n}
\DeclareMathSymbol{\VDash}{\mathop}{MnSyA}{240}
\DeclareMathSymbol{\downsquig}{\mathop}{MnSyA}{163}
\DeclareMathSymbol{\rightsquig}{\mathop}{MnSyA}{160}

\usepackage{refcount}
\newcounter{fncntr}
\newcommand{\fnmark}[1]{\refstepcounter{fncntr}\label{#1}\footnotemark[\getrefnumber{#1}]}
\newcommand{\fntext}[2]{\footnotetext[\getrefnumber{#1}]{#2}}

\newcommand{\mydots}{\ifmmode\mathinner{\kern-0.1em\ldotp\kern-0.1em\ldotp\kern-0.1em\ldotp\kern-0.1em}\else.\kern-0.13em.\kern-0.13em.\fi}
\newcommand{\jmydots}{\ifmmode\mathinner{\kern-0.2em\ldotp\kern-0.2em\ldotp\kern-0.2em\ldotp\kern-0.2em}\else.\kern-0.2em.\kern-0.2em.\fi}
\newcommand{\Bdots}{\ifmmode\mathinner{\kern1mm\ldotp\kern2mm\ldotp\kern2mm\ldotp\kern1mm}\else.\kern-0.13em.\kern-0.13em.\fi}
\newcommand{\srcsize}{\@setfontsize{\srcsize}{4pt}{4pt}}

\begin{document}

\AuthorTitle{{\large This is the Author's Manuscript for a forthcoming article in \textit{Logic \& Logical Philosophy}. The final authenticated version is DOI: 10.12775/LLP.2020.019}\\\vspace{1cm}Jared Millson}{A Defeasible Calculus for Zetetic Agents}



\allowdisplaybreaks
\sloppy

\Abstract{The study of defeasible reasoning unites epistemologists with those working in AI, in part, because both are interested in epistemic rationality. While it is traditionally thought to govern the formation and (with)holding of beliefs, epistemic rationality may also apply to the interrogative attitudes associated with our core epistemic practice of inquiry, such as wondering, investigating, and curiosity. Since generally intelligent systems should be capable of rational inquiry, AI researchers have a natural interest in the norms that govern interrogative attitudes. Following its recent coinage, we use the term ``zetetic'' to refer to the properties and norms associated with the capacity to inquire. In this paper, we argue that zetetic norms can be modeled via defeasible inferences to and from questions---a.k.a \textit{erotetic} inferences---in a manner similar to the way norms of epistemic rationality are represented by defeasible inference rules. We offer a sequent calculus that accommodates the unique features of ``erotetic defeat"  and that exhibits the computational properties needed to inform the design of zetetic agents. The calculus presented here is an improved version of the one presented in \cite{Millson2019}, extended to cover a new class of defeasible erotetic inferences.}


\section{Agents, Rationality, and Inquiry}
The study and modeling of defeasible reasoning unites epistemologists with those working in artificial intelligence (AI). This convergence is due, in part, to a joint interest in epistemic rationality. Traditionally, epistemically rational behavior is conceived in terms of conformity with rules, standards, or ideals specifying the conditions under which an agent is permitted (or obliged) to form or withhold beliefs. Since these norms are commonly understood as admitting exceptions or conflicts, the concept of defeasibility has occupied an important role in their study. Epistemic rationality is thus often identified with the defeasible inferences an agent is (epistemically) permitted to draw. Naturally, this is especially relevant to work in AI, where the goal is, \textit{inter alia}, the construction of artificial agents capable of operating with a range of information sources in an epistemically rational manner.  

It would be a mistake, however, to think that epistemic rationality only governs the formation or withholding of beliefs. In recent years, epistemologists and philosophers of science have come to recognize an array of non-doxastic or quasi-doxastic states and processes that are subject to epistemic norms \citep{Fleisher2018,Friedman2013a,Lacey2015,Whitcomb2017,Palmira2019}. These include, most notably, the attitudes associated with our core epistemic practice of \textit{inquiry}.

An inquiring agent aims to improve its epistemic standing with respect to some subject-matter---i.e. to obtain, in some sense, the correct answer to a question---and undertakes certain actions in order to do so, e.g. gathering evidence. But since it is possible to perform those same actions so as to mislead others into thinking that one is inquiring, these actions cannot be sufficient. To qualify as genuinely inquiring into an issue, an agent must have an \textit{interrogative attitude} that is oriented toward settling the question at hand and that is sensitive to information bearing on it \citep{Friedman2013a}. In contrast with propositional attitudes like belief, the contents of these attitudes are questions.  Interrogative attitudes include \textit{wondering}, \textit{being curious}, \textit{investigating}, \textit{contemplating}, \textit{deliberating}, and \textit{suspending judgment} about $ Q $. In what follows, we use the verb-phrase ``inquiring into $ Q $'' as a place-holder for any one of these attitudes. One way to think about the epistemic norms that govern inquiry, then, is to consider when it is epistemically rational for an agent to adopt an interrogative attitude toward a question, i.e. to inquire into $ Q $. Following \cite{Friedman2019a}, we'll refer to the norms that govern these attitudes and to the rationality they embody as \textit{zetetic} from the ancient Greek word for inquiry, \textit{z\^{e}t\^{e}sis}. 

Unfortunately, this new line of research has yet to make full contact with the field of AI, despite the latter's obvious interest in developing systems that perform information-seeking operations. Intelligent agents are able, e.g., to search for relevant information at appropriate intervals and to `raise' or `drop' questions in response to changes in information states. Behaviors such as these may serve as an intermediate step toward an agent's goal of maximizing some external reward. But inquisitive behavior and the states underlying it can play other roles as well. For instance, in environments where extrinsic rewards are scarce, interrogative attitudes such as \textit{curiosity} can serve as intrinsic reward signals, encouraging the agent to explore its environment, develop successful strategies, and learn useful skills \citep{Burda2018,Graziano2011,Pathak2017,Pape2012}. 

Among the areas where such artificially curious agents might have a significant impact is in the automation of scientific discovery \citep{Savage2012}. Autonomous agents capable of scientific activity will need to be able to determine which lines of inquiry are worth pursuing. So, automating these decisions via artificial curiosity is likely to be central to the project of automating scientific activity in general. Extending the study of epistemic rationality to the behavior and attitudes associated with inquiry is therefore important for epistemologists and AI researchers alike.

Typically, defeasible inferences are understood along the lines of the ``If-Then-Unless'' pattern, where conclusions (the `Then' part) are drawn `usually', `normally', or `by default' when certain information is present (the `If' part) but are withheld, relinquished, or bracketed when information defeating that inference is detected (the `Unless' part).\fnmark{def}  An agent who draws such inferences in the presence of defeaters exhibits, \textit{ceteris paribus}, a sort of incoherence, such as that associated with having inconsistent beliefs or adopting incompatible doxastic attitudes toward the same content. If epistemic rationality is generally associated with defeasible inferences, then we would expect zetetic rationality to be similarly defeasible. But what collateral states would make it irrational or incoherent for an agent to adopt an interrogative attitude like \textit{wondering whether} $ p $?

\fntext{def}{See \cite{Bloeser2013} for a nice discussion of the extensive literature on defeasibility across logic, epistemology, and law.}

Linguistic practice provides a clue. A speaker who asks a question often flouts conversational maxims when it is common knowledge that it knows or fully believes the answer.\fnmark{test} Of course, it is often quite useful and even necessary to flout such maxims in order to achieve certain communicative goals, e.g. to feign ignorance. But the dialectical proscription on asking information-seeking questions for which one has the answer may reveal a distinctly epistemic norm. The suggestion that has recently emerged is that it is epistemically irrational, at least \textit{prima facie}, to adopt interrogative attitudes toward questions whose answer is (already) believed or known. Stated more carefully, an agent's epistemic license or entitlement to inquire into a question is defeated when it knows or is licensed to believe its answer \citep{Friedman2017,Friedman2017a,Whitcomb2017}. To give zetetic rationality its due, this \textit{erotetic defeat} needs to be investigated alongside the kinds of defeat that are traditionally represented in nonomonotonic logics.

\fntext{test}{\hspace{1mm}One obvious exception is when a speaker asks a so-called ``exam question'' and intends to test the addressee's knowledge.}

The goal of this paper is to lay the ground for a formal study of zetetic rationality that is of recognizable import to epistemologists---at least to those who work in the formal and logical areas of that field---as well as to AI researchers. To do so, we examine defeasible inferences involving the contents of interrogative attitudes, i.e. questions, in addition to the contents of doxastic attitudes, i.e. propositions. The former have traditionally been the subject matter of erotetic logic. What we propose, then, is a defeasible erotetic logic with the computational properties needed to inform the design of zetetically intelligent systems. 

We draw on work in \textsc{Inferential Erotetic Logic} (IEL), as it provides a familiar truth-conditional semantics for erotetic inferences and identifies some of them as nonmonotonic \citep{Wisniewski1995,Wisniewski2013}. While the relations between questions and answers have received various logical treatments \citep{Belnap1976,Hamblin1958,Hintikka2007,Ciardelli2018}, IEL is rather unique in its focus on inferences among questions themselves. IEL's approach is to start with a standard propositional language and to enrich it syntactically with question-forming vocabulary. Following the \textit{set-of-answers} methodology, IEL represents questions as sets of their possible direct answers, with the latter being construed as syntactically distinct, well-formed formulas (wffs) of the propositional language. In virtue of this capacious construal of questions, IEL can readily accommodate the popular intensional semantics for natural language interrogatives that treats questions (i.e. the content of interrogative clauses) as partitions of logical space \citep{Groenendijk1996}. In recent years, the logical study of questions has benefited from the rise of inquisitive semantics, a program which, unlike IEL, does not assume a syntactic distinction between declaratives and interrogatives but rather provides a unified semantic account according to which informative content may be distinguished from inquisitive content. Fortunately, many of the relations IEL studies have known correlates in inquisitive semantics \citep{Winiewski2015}.

What we want is a calculus that facilitates (re)construction of reasoning in which interrogative attitudes are adopted and abandoned. IEL is a good resource for this, given that some of its relations are nonmonotonic. Unfortunately, there has been very little proof theoretic work done with erotetic logics, let alone IEl. The few calculi that exist for species of IEL's relations either do not encode defeasibility \citep{Wisniewski2016,Wisniewski2004,Leszczyska-Jasion2008,Leszczynska-Jasion2013}, or do not capture the type of defeasible erotetic inferences that take questions as both premises and conclusion \citep{Meheus1999,Meheus2001,Millson2019}. In fact, the calculus presented in \cite{Millson2019}, $ \mathsf{LK^?} $, cannot capture these inferences at all. The limitation  arises from the fact that sequents in $ \mathsf{LK^?} $ contain an auxiliary syntactic parameter---so-called ``Background Sets''---that preserve a `trace' of active formulas that are otherwise eliminated as the proof proceeds downward. Although they greatly aid the proof of a cut-elimination theorem, these background sets prevent the representation of erotetic defeat in cases of question-to-question inferences and produce a significant syntactic `bloat,' since the size of background sets increases dramatically through the course of a proof.\fnmark{lemma3} The system presented here does away with the background-set device and offers a far more streamlined and elegant calculus---one that is shown to be sound and complete without the use of a cut rule.

\fntext{lemma3}{In \cite{Millson2019}, the background-set device facilitates the proof of the property established in Lemma \ref{lem:Derivable_undefeated_provable_SCs} below.}

Filling this proof-theroetic lacuna is a central aim of the present work. We offer a decidable sequent calculus that is sound and complete for IEL's defeasible erotetic-inferential relations. The calculus includes introduction and elimination rules for interrogative-forming expressions that correspond, intuitively, to the adoption of interrogative attitudes. We hope, then, to have an (informally) well-founded and (formally) well-behaved calculus for studying zetetically rational agents.

The system presented here might aid the engineering and design of such agents. For instance, if the logic is decidable for proofs of defeasible inferences between questions, then an automated theorem prover (ATP) could help agents determine which auxiliary or subordinate questions they are epistemically permitted to inquire into given their initial question(s) and knowledge base. Since the inferences are erotetically defeasible, the ATP could also be used to tell the agent when it has obtained an answer to one of its questions and thus is no longer licensed to inquire. Such an ATP could nicely complement the reinforcement learning of artificially curious agents.

The paper proceeds by introducing IEL's semantic definitions for two types of erotetic inference (\S \ref{sec:IEL}). Next, we define a sequent calculus for the language of classical propositional logic (CPL) that tracks the defeat of sequents on the basis of a finite set of defeaters assigned to each atom by the axioms (\S \ref{subsec:SCS?}). We then extend the calculus with rules for erotetic formulas (\S \ref{subsec:SCp?rules}) and show that the calculus is decidable and is sound and complete for IEL's defeasible erotetic inference-relations (\S \ref{sec:Results}). We finish with a worked example (\S \ref{sec:ex}) and some suggestions for future development (\S \ref{sec:concl}).

\section{Erotetic Inferences}\label{sec:IEL}

Inferential Erotetic Logic (IEL) studies the relations underwriting two patterns of reasoning in which questions play a crucial role. The first occurs when information `opens up' or makes salient a set of possibilities, each of which provides a possible answer to a particular question. In these situations, we might say that our state of information, set of beliefs, or theoretical commitments gives rise to or \textit{evokes} a question. For example, knowing that someone broke the vase may give rise to the question of who, among a certain set of individuals, broke it. It's natural to think of our reasoning here as one of inferring questions from statements. But we also infer questions from other questions. We may find that progress can be made in answering one question by answering another. If we are trying to figure out who broke the vase and we have evidence that whoever broke it is wearing sandals, we can narrow down the set of potential culprits by asking who of them is wearing sandals. Here, our initial question, together with our evidence regarding the culprit's footwear, form the premises from which we infer the latter question. The first of these patterns is called question or erotetic \textit{evocation}, the second, \textit{erotetic implication} (hereafter: e-implication).

Evocation and e-implication specify conditions under which an agent is entitled to ask or inquire into a question, in the sense of adopting an interrogative attitude. By contrast, erotetic defeat describes a situation in which an agent is not licensed or is forbidden to inquire into a question. So evocation and e-implication are undermined when an agent has an answer to the relevant questions. In drawing one of these types of inference, agents reserve the right, so to speak, to retract or withhold their entitlement to inquire in the light of further evidence, i.e. if they obtain the answers to their questions. Evocation, for instance, fails to obtain when the declarative premises entail an answer to the evoked question. Similarly, e-implication may fail when the premises entail an answer to the initial (i.e. implying) question. Since an agent may deploy these inference-patterns only insofar as it doesn't already have the answers to its questions, IEL provides inferential relations that conform to our intuitions about erotetic defeat.

Initially  proposed by Andrzej Wi\'{s}niewski, \textsc{Inferential Erotetic Logic} (IEL) offers a generalized semantic account of evocation and e-implication. Since our aim is to develop a proof theory for these inferences with the prospects of implementation---.e.g. in the form of an automated theorem prover---we will be concerned with the expression of these inferences in a familiar logical language, namely, that of classical propositional logic (CPL). We focus on what is known in IEL as ``regular" e-implication, which is transitive. Doing so allows us to formulate a calculus that can be equipped with a cut rule that applies to either d-wffs or e-wffs. In the conclusion, we propose an extension of the calculus that covers these non-regular inferences.

Let $ \mathcal{L} $ be a formal language composed of two disjoint fragments. The first, $ \mathcal{L}_d $, consists of declarative well-formed formulas (hereafter: `d-wffs' or `statements'); the second, $ \mathcal{L}_e $, consists of interrogative formulas (hereafter: `e-wffs' or `questions'). The ``declarative part'' of $ \mathcal{L} $, i.e. $ \mathcal{L}_d $, is the language of classical propositional logic (CPL) with a countable set of atoms, $ \mathcal{A} $, which are denoted $p,q,r, s, t, u, v$, sometimes with subscripts.  Arbitrary d-wffs are represented by $A, B, C, D$, (possibly empty) sets of d-wffs by $ X, \varUpsilon, Z $, and (possibly empty) sets of sets of d-wffs by $ \mathbf{S}, \mathbf{T}, \mathbf{U}, \mathbf{V}$---each may be indexed with the help of subscripts. 	

The ``erotetic'' part of $ \mathcal{L} $, is formed by enriching $ \mathcal{L}_d $ with ``?,'' ``\{ ,'' ``\},'' and the comma---thus yielding e-wffs such as $ ?\{p\wedge q, r\vee q, \neg p \} $.  The metavariables $ Q, R $ range over e-wffs. We refer to the d-wffs from which questions are formed as their \textit{direct answers} and will use $ d(\cdot) $ to denote the function that maps questions to the set of their direct answers. So, when $Q = ?\{p, \neg q \} $, $ dQ = \{p, \neg q\} $. We stipulate that $ \mathsf{card}(dQ)\geq 2 $ for any $ Q\in\mathcal{L}$. The metavariables  $ F, G, H $ range over formulas and $ \Gamma, \Delta $ over sets of formulas of either type (i.e. statements or questions) in $ \mathcal{L} $. We will use `$ = $' to denote both set-theoretic identity as well as syntactic equivalence.

\begin{definition}[Syntax of $ \mathcal{L} $]\label{def:syntax_L}
	$ \mathcal{L} $ is the smallest set meeting the following criteria:
	\begin{itemize}
		
		\item[(i)] If $ A \in \mathcal{L}_d$, then $ A \in \mathcal{L} $.
		
		\item[(ii)] If	$ A_{1},\ldots,A_{n}  (n >1) \in \mathcal{L}_d$ and $  A_{1},\ldots,A_{n}$ are pairwise syntactically distinct, i.e., non-equiform, then $ ?\{A_{1},\ldots,A_{n}\} \in \mathcal{L}$.
		
		\item[(iii)] Nothing else is a member of $ \mathcal{L}$.
		
	\end{itemize}
\end{definition}

In their most general form, evocation and e-implication are explicated in terms of two entailment relations, one for single-conclusions and another for multiple conclusions. But since we will be working in the restricted context of CPL, we only need to use the standard, single-conclusion entailment relation for classical logic, denoted by `$ \vDash $.' The reader is welcome to supply this relation with her preferred semantics for CPL.

\begin{definition}[Evocation]\label{def:erotetic_evocation}
	If set of d-wffs, $ X $, evokes a question, $ Q $, we write $ X\rightarrowtail Q  $.
	\begin{equation}\nonumber
	X\rightarrowtail Q \textit{\:\:iff\:\:}
	\begin{cases}
	\text{(i)} & X \vDash \bigvee dQ {\:\:but} \\
	\text{(ii)} & \forall A \in dQ (X \nvDash A)
	\end{cases}
	\end{equation}
\end{definition}

The first condition for erotetic evocation requires that an evoked question, $ Q $, has a true direct answer if the evoking set consists of truths. A question is said to be \textit{sound} iff at least one direct answer to the question is true. So the first clause of Definition \ref{def:erotetic_evocation} amounts to transmission of truth to soundness. 

The second condition says that no single direct answer to $ Q $ is entailed by $ X $. This constraint encodes the concept of erotetic defeat---i.e. when an agent's information state (already) entails a direct answer to a question, it is not entitled to inquire into it. Since it is possible for  $ X\rightarrowtail Q $ to hold while for some $ X'\supset X $, $ X'\not\!\rightarrowtail Q $, the evocation relation is not monotonic. 

IEL describes a variety of e-implication relations, but for our purposes, we want to focus on one that is both transitive---so that we may have a rule of \textit{cut} in our system---and defeasible. In the language of IEL, such a relation is \textit{strong}, in the sense of being defeasible, and \textit{regular} in the sense that each answer to the implied question entails an answer to the implying one, rather than merely entailing a proper subset of answers as general e-implication does. 

\begin{definition}[Strong Regular Erotetic Implication, e$_{sr}$-implication]\label{def:sr_e-imp}
	If a question $ Q $ strongly regularly implies a question $ Q' $ on the basis of a set of d-wffs $ X $, we write $ X\mid Q \twoheadrightarrow Q' $.
	\begin{equation}\nonumber
	X\mid Q \twoheadrightarrow Q' \textit{\:\:iff\:\:}
	\begin{cases}
	\text{(i)} & \forall A \in dQ\: (X \cup \{A\} \vDash \bigvee dQ' ) {\:\:and} \\
	\text{(ii)} & \forall B \in dQ'\:\exists A \in dQ \,(X \cup \{B\} \vDash A) {\:\:but}\\
	\text{(iii)} & \forall A \in dQ\: (X \nvDash A)
	\end{cases}
	\end{equation}
\end{definition}

The first clause in Definition \ref{def:sr_e-imp} says that the implying question, together with a set of auxiliary d-wffs, $ X $, entails that at least one direct answers to the implied question is true. This ensures that the relation transmits the soundness of $ Q $ and the truth of $ X $ to the soundness of $ Q' $. The second clause guarantees that $ Q' $ is \textit{cognitively useful} relative to $ X $ and $ Q $, in the sense that each direct answer to the implied question, together with the auxiliary d-wffs, provides an answer to the initial, implying question. (For the general relation of e-implication, cognitive utility need only consist in \textit{narrowing down} the set of direct answers to the implying question). This condition aims to redeem the intuition that pursuing an implied question is an epistemically rational way of inquiring into the implying question---i.e. that drawing inferences in conformity with e-implication is a rational way of making progress in inquiry. The third clause in Definition \ref{def:sr_e-imp} ``strengthens'' regular e-implication into a nonmonontonic relation by encoding the conditions of erotetic defeat. It tells us that the erotetic inference is defeated when an agent is (already) entitled to believe an answer to its implying question.

The statements of defeasibility in these definitions are quite similar. We may consider both evocation and strong regular e-implication (hereafter: e$_{sr}$-implication) as specifying conditions under which an agent is entitled to inquire into a question that forms the conclusion of an inference. Definition \ref{def:erotetic_evocation}(ii) tells us that this entitlement is lost if an agent has or acquires information that licenses a belief in one of its answers. In other words, agents should not inquire into questions for which they have the answer. Definition \ref{def:sr_e-imp}(iii) states that this entitlement is lost if an agent has or acquires information that licenses it to believe the initial, implying question's answer. The entitlement to inquire into the implied question, $ Q' $, is, after all, based on its entitlement to inquire into the initial question, $ Q $---i.e. inquiry into $ Q' $ is a kind of rational strategy for pursing inquiry into $ Q $. Thus, if an agent has the answer to $ Q $ there ceases to be a reason, \textit{ceteris paribus}, to inquire into $ Q' $. But if condition (ii) in Definition \ref{def:sr_e-imp} holds, then e$_{sr}$-implication will be similarly defeated when an agent has an answer to the implied question, $ Q' $, since if $ X\vDash B $ for some $ B\in dQ' $ and if $ \forall B \in dQ', X \cup \{B\} \vDash A $ for some $ A\in dQ  $, then $ X\vDash A $, thus violating (iii). Evocation and e$_{sr}$-implication are both therefore defeated when the declarative premises entail answers to their question-conclusions.

Some other features of IEL's representation of erotetic defeat are worth considering. Both inferences are defeated, for instance, when their declarative premises are inconsistent. If $ X\vDash $, then $ X\not\!\rightarrowtail Q $ and $ X\mid Q \not\twoheadrightarrow Q' $, since $ X\vDash A $ for any $ A \in dQ $. This result obtains because of CPL's respect for the principle of explosion or \textit{ex contradictione quod libet}.  Philosophers, logicians, and AI researchers have long criticized the principle of explosion as unrepresentative of ordinary reasoning and have promoted paraconsistent systems in which it does not hold. But the behavior of evocation and e$_{sr}$-implication here appears to capture an important kind of incoherence that characterizes agents trying to conduct inquiry in the midst of inconsistent information. The calculus we present below respects this fact and treats an inference with inconsistent premises, i.e. any sequent with an inconsistent antecedent, as defeated. Thus, technically speaking, the system is paraconsistent insofar as there is no provable sequent corresponding to the principle of \textit{ex contradictione quod libet}.

Both evocations and e$_{sr}$-implications are also defeated when one of their questions has tautologous direct answers, i.e. if some $ A\in dQ $ is valid in CPL. If there is some $ A \in dQ$ such that $ \vDash A $, then $ X\not\!\rightarrowtail Q $ and $ X\mid Q \not\twoheadrightarrow Q' $, since $ X\vDash A $. Concerns about this behavior may be raised on the grounds that it encodes an unrealistic assumption of logical omniscience. Even highly rational agents often fail to recognize the logical consequences of their beliefs, let alone tautologies. So, it seems equally unrealistic to model agents as infallible detectors of logical truths. But the defeat of these inferences in cases where their question have tautological answers aligns with the intuition that questions with such non-factual answers are often redundant, pointless, or otherwise fail to facilitate rational inquiry, at least that which deals with empirical subject matter. Again, exploring alternative formulations of these inferential relations that do not behave in this manner may be worthwhile.

\section{The Calculus $ \mathsf{SC}^? $}\label{sec:SCp?}

Now that we've seen the kinds of defeasible relations that IEL has to offer, the challenge is to devise a proof theory capable of encoding these relations as inference rules. Since we want our calculus to incorporate defeasible inferences involving both declarative and erotetic formulas, and since each rule will either introduce or eliminate a connective, we will have distinct rules for propositional and erotetic connectives, with the latter confined to interrogative-forming expressions of the form ``$ ?\{\ldots\} $''. We begin by formulating the calculus for d-wffs, $\mathsf{SC}^\mathbf{S} $ and proceed by extending it with rules for e-wffs.

\subsection{The propositional sequent calculus $\mathsf{SC}^\mathbf{S} $}\label{subsec:SCS?}

Our calculus for defeasible propositional inferences, $\mathsf{SC}^\mathbf{S} $, uses standard multiple-succedent sequents composed of \textit{sets} of formulas, e.g., $\Gamma\vdash\Delta $. Formulas on the left side of the turnstile are called the \textit{antecedent}; on the right side they are called the \textit{succedent}. Commas in the antecedent are read conjunctively and those in the succedent are read disjunctively. The formula with the connective in a rule is the \textit{principal} formula of that rule, and its components in the premises are the \textit{active} formulas. The remaining elements of sequents are referred to as \textit{side formulas}. But unlike standard sequents, ours include a set of sets of d-wffs, called a \textit{defeater set}, just below our turnstile, e.g. $ \Gamma\sststile{\mathbf{S}}{}\Delta $.\fnmark{Piazza}

\fntext{Piazza}{The syntax of $\mathsf{SC}^\mathbf{S} $ sequents is inspired by the calculus in \cite{Piazza2017}. }

We interpret sequents as representing a relationship that preserves an agent's epistemic entitlement to adopt doxastic and interrogative attitudes. We use the terms `entitled,' `licensed,' `rational,' and `permissible' interchangeably to denote this preserved status. Thus, a provable sequent says that any agent entitled to \textit{believe} the statements and to \textit{inquire into} the questions in the antecedent is entitled to \textit{believe} or \textit{inquire into} at least one of the statements/questions in the succedent. While sequents represent rules of entitlement-preserving inference, the rules of our calculus are meta-rules of inference---i.e., they tell us what rules of inference are legitimate given the axioms. 

Defeater sets represent situations in which one is not permitted to draw the inference represented by the sequent. A purely declarative sequent such as $ A\sststile{\{C_1, C_2\}, \{C_3\}}{}B $ says that an agent who is entitled to believe that $ A $ is entitled to believe that $ B $ unless it is entitled to believe $ C_1 $ or $ C_2 $ or to believe that $ C_3 $. 

As this interpretation suggests, a sequent is defeated when an agent is licensed to believe at least one statement in a member (set) of the defeater set. (As we explain in Section \ref{subsec:ED}, treating defeater sets as sets of sets permits us to represent the peculiar conditions of erotetic defeat.) D-wffs in the antecedent of a sequent specify which statements an agent is entitled to believe. Roughly put, a sequent is defeated when some statement in a member (set) of the defeater set is the (classical) logical consequence of its antecedent. This amounts to assuming that an agent's belief set is closed under (classical) logical consequence.

$\mathsf{SC}^\mathbf{S} $ is inspired by Smullyan's \citeyearpar{Smullyan1968} ``symmetric'' sequent calculus for CPL, $ \mathscr{S} $, though it departs from that calculus in certain respects, canvassed below. The core characteristic of $ \mathscr{S} $ that is carried over to $\mathsf{SC}^\mathbf{S} $ is that the calculus does not allow for switching formulas from one side of a sequent to another. Consequently, an end-sequent's subformulas occur on the same side of the turnstile throughout a derivation. As we shall see, this constraint is important for preserving the intuitive meaning of defeasible sequents. The rules for $\mathsf{SC}^\mathbf{S} $ are given in Figure \ref{fig:SCd}. Note that sets of side formulas that appear in these rules are represented by $\Gamma$ and $ \Delta $ which we defined as ranging over members of $ \mathcal{P}(\mathcal{L}).$ Strictly speaking, however, $\mathsf{SC}^\mathbf{S} $ is defined over $ \mathcal{L}_d $. We use the metavariables $ \Gamma, \Delta $ to obviate re-writing the rules when we extend the calculus for $ \mathcal{L} $.

Like Smullyan's $ \mathscr{S} $, our calculus, $\mathsf{SC}^\mathbf{S} $, replaces the familiar negation rules of $ \mathsf{LK} $ with rules for double negation and De Morgan's Laws, as well as two additional axioms. But in contrast with $ \mathscr{S} $, $\mathsf{SC}^\mathbf{S} $ includes structural rules, multiplicative rather than additive versions of the rules for connectives, and axioms (four rather than three as in $ \mathscr{S} $) that do not permit arbitrary side-formulas. Proofs in the calculus are the familiar tree structures of sequent calculi, but since sequents may be defeated, we distinguish between derivations (i.e. trees constructed via the application of rules irrespective of whether the sequents are defeated) and proofs (i.e. trees in which no defeated sequents occur). 

\begin{definition}[Derivation, Proof, Paraproof]\label{def:proof}
	If a defeasible sequent, $\Gamma\sststile{\mathbf{S}}{}\Delta $, occurs at the root of a finitely branching tree, $ \pi $, whose nodes are defeasible sequents recursively built up from axioms by means of the rules of $ \mathsf{SC}^\mathbf{S}$ ($ \mathsf{SC}^?$), then $ \pi $ is said to be a \textit{derivation} of $ \Gamma\sststile{\mathbf{S}}{}\Delta $. If, additionally, each sequent in $ \pi $ is undefeated, then $ \pi $ is said to be a \textit{proof} of $ \Gamma\sststile{\mathbf{S}}{}\Delta $ in $\mathsf{SC}^\mathbf{S}$ ($ \mathsf{SC}^?$),\, otherwise $ \pi $ is called a \textit{paraproof} of $\Gamma\sststile{\mathbf{S}}{}\Delta $.
\end{definition}

So defeat and defeasibility play no role in determining which sequents are derivable in $\mathsf{SC}^\mathbf{S} $. Indeed, if we restrict our attention to derivable sequents in $\mathsf{SC}^\mathbf{S} $, we have a calculus that is but a variation of Smullyan's  $ \mathscr{S} $---differing in the aspects noted above.

\begin{lemma}\label{lem:SCtoSCS}
	$ X\sststile{\mathbf{S}}{}\varUpsilon$ is derivable in  $\mathsf{SC}^\mathbf{S} $ iff $ X\vdash\varUpsilon $ is provable in $ \mathscr{S} $.
\end{lemma}

Smullyan's $ \mathscr{S} $ is sound and complete for CPL. So, it follows from Lemma \ref{lem:SCtoSCS} that the antecedent of any  sequent derivable in  $\mathsf{SC}^\mathbf{S} $ classically entails the disjunction of the elements in the succedent.

\begin{lemma}\label{lem:SCS=CPL}
	$ X\sststile{\mathbf{S}}{}\varUpsilon$ is derivable in $\mathsf{SC}^\mathbf{S} $ iff $ X\vDash\bigvee\varUpsilon $.
\end{lemma}

Derivations in the system therefore constitute a calculus that is sound and complete for CPL. The upshot of the distinction between derivations and proofs in $\mathsf{SC}^\mathbf{S} $ is that we are able to mobilize the classical consequence relation without having to appeal to an ``external'' calculus. This, in turn, permits us to define the conditions of defeat internally, making $\mathsf{SC}^\mathbf{S} $, in a sense, self-contained.

\begin{definition}[Defeat in $\mathsf{SC}^\mathbf{S} $]\label{def:defeat_in_SCd} A sequent is defeated just in case there is sequent derivable in $\mathsf{SC}^\mathbf{S} $ that has the same antecedent and a succedent that is a member of the original sequent's defeater set. (The content of the defeater set of this derivable sequent, i.e. $\mathbf{T}$, is irrelevant.)		
	$$   X\sststile{\mathbf{S}}{}\varUpsilon\text{ \: is defeated \textit{iff} \:} \exists Z\in\mathbf{S} \text{\, such that \,} X\sststile{\mathbf{T}}{}Z \text{\:is derivable in \:}\mathsf{SC}^\mathbf{S}. $$
\end{definition}

The definition of \textit{defeat} is designed to redeem certain intuitions about defeasible inferences. For instance, if $ A $ defeats an inference, then an agent entitled to $ A\wedge B $ ought not to make it (Example \ref{ex:con_in_ant}).  Conversely, if $ A \wedge B $ defeats an inference but neither $ A $ nor $ B $ does by itself, then the agent entitled to only one of the conjuncts may still draw the inference (Example \ref{ex:con_in_def}). Similarly, if $ A\vee B $ defeats an inference, then the inference is defeated if the agent is entitled to either of the disjuncts (Example \ref{ex:dis_in_def}). Finally, if $ A $ defeats an inference, then entitlement to $ A\vee B $ need not force an agent to reject it, since entitlement to $ B $ would preserve the propriety of the inference (Example \ref{ex:dis_in_ant}).

\begin{example}\label{ex:con_in_ant}
	$ p\wedge q \sststile{ \{\{p \}\} }{} \Delta $ is defeated.
\end{example}

\begin{example}\label{ex:con_in_def}
	$ p\sststile{\{\{p \wedge q\}\}}{} \Delta $ is undefeated.
\end{example}

\begin{example}\label{ex:dis_in_def}
	$ p \sststile{\{\{p\vee q \}\}}{} \Delta $ is defeated.
\end{example}

\begin{example}\label{ex:dis_in_ant}
	$ p \vee q \sststile{\{\{p\}\}}{} \Delta $ is undefeated.
\end{example}

Notice that when a sequent has an empty defeater set, i.e. $ \Gamma\sststile{\emptyset}{}\Delta $, the defeat mechanism in $\mathsf{SC}^\mathbf{S} $ is idle. Typically, defeasible logics contain defeasible or ``non-strict'' rules and indefeasible or ``strict'' rules. In $\mathsf{SC}^\mathbf{S} $ strict inference rules are represented by these sequents with empty defeater sets, while defeasible inferences are captured by sequents with nonempty defeater sets. (Since axioms may have empty or nonempty defeater sets, the calculus permits a mix of strict and non-strict rules.) 

With the notion of defeat in hand, we can now observe the advantage accrued by using ``symmetric'' sequent rules---namely, to remain faithful to the informal reading of sequents as ordinary defeasible inferences and, in keeping with that goal, to ensure that unary (i.e. one-premise) rules preserve defeat downward and undefeatedness upward.

\begin{fact}[Unary Rules Preserve Defeat Downward]\label{fact:DdefeatUnary}
	If $ \Gamma\sststile{\mathbf{S}}{}\Delta $ follows from $ \Gamma'\sststile{\mathbf{S'}}{}\Delta' $ via a unary rule (i.e. $\mathsf{LW},\mathsf{RW},\mathsf{DE},{\wedge}{\vdash}, {\vdash}{\vee}, {\neg}{\neg}{\vdash}, {\vdash}{\neg}{\neg},\\ {\vdash}{\neg}{\wedge}, {\neg}{\vee}{\vdash} $), then $ \Gamma\sststile{\mathbf{S}}{}\Delta $ is defeated if $ \Gamma'\sststile{\mathbf{S'}}{}\Delta' $ is defeated. Contrapositively, if $ \Gamma\sststile{\mathbf{S}}{}\Delta $ is undefeated, then $ \Gamma'\sststile{\mathbf{S'}}{}\Delta' $ is undefeated.
\end{fact}

\noindent The right rule for negation in $ \mathsf{LK} $, on the other hand, frustrates both desiderata.
$$\begin{prooftree}
\Hypo{\Gamma,A\sststile{\mathbf{S}}{}\Delta}
\Infer1[$\vdash\neg$]{\Gamma\sststile{\mathbf{S}}{}\neg A, \Delta}
\end{prooftree}$$
\noindent If, e.g., $ A \in \bigcup\mathbf{S} $, then the upper sequent is defeated while the lower sequent is not. It is rather counter-intuitive to think that applying a purely logical, unary rule of meta-inference should have this effect. Moreover, it is quite difficult to give an intuitively plausible reading of this rule as it appears to warrant an agent's abandoning entitlement to a belief, i.e. that $ A $. 

The binary rules ${\vdash}{\wedge}$ and ${\vdash}{\neg\vee}$ also preserve defeat downwards and undefeatedness upwards. They share this property in virtue of their multiplicative formulation. Since the antecedents of premises-sequents in this rule are combined in the conclusion, any defeat that occurs in the premises is carried downward. 

\begin{fact}[${\vdash}{\wedge}$ and ${\vdash}{\neg\vee}$ Preserve Defeat Downward]\label{fact:Ddefeattwobinary}
	If $ \Gamma\sststile{\mathbf{S}}{}\Delta $ follows from $ \Gamma'\sststile{\mathbf{S'}}{}\Delta' $ and $ \Gamma''\sststile{\mathbf{S''}}{}\Delta'' $ via ${\vdash}{\wedge}$ or ${\vdash}{\neg\vee}$, then $ \Gamma\sststile{\mathbf{S}}{}\Delta $ is defeated if either $ \Gamma'\sststile{\mathbf{S'}}{}\Delta' $ or $ \Gamma''\sststile{\mathbf{S''}}{}\Delta'' $ is defeated. Contrapositively, if $ \Gamma\sststile{\mathbf{S}}{}\Delta $ is undefeated, then $ \Gamma'\sststile{\mathbf{S'}}{}\Delta' $ and $ \Gamma''\sststile{\mathbf{S''}}{}\Delta'' $ are undefeated.
\end{fact}

In contrast, the binary rules $ {\vee}{\vdash} $ and $ {\neg}{\wedge}{\vdash} $ permit the derivation of undefeated sequents from defeated ones. This fact, which figures prominently in the proof of Lemma \ref{lem:Derivable_undefeated_provable_SCs} below, arises because the rules' active formulas are logically stronger than their principal formula. Consequently, a member of the conclusion's defeater set may follow (classically) from either of the premises' antecedents while not following from the conclusion's. 

\begin{fact}[Derivation of Undefeated Sequents from Defeated Ones]\label{fact:DeriUNDefeat}
	Undefeated sequents can only be derived from (one or more) defeated sequents via $ {\vee}{\vdash} $ and $ {\neg}{\wedge}{\vdash} $. 
\end{fact}

The inclusion in $\mathsf{SC}^\mathbf{S} $ of a rule for left weakening may surprise the reader, as this rule is typically associated with monotonicity. When a sequent's antecedent is weakened by a formula that entails a member of the defeater set, however, the result is a defeated sequent. As Definition \ref{def:proof} makes clear, proofs in the calculus must not contain defeated sequents. Thus, provability in the calculus will indeed conform to nonmonotonicity. A rule for expanding the defeater set, $ \mathsf{DE} $, is also provided in order to represent cases of reasoning in which new information about defeaters is acquired.

While undefeated sequents can only be obtained from defeated ones via $ {\vee}{\vdash} $ or $ {\neg}{\wedge}{\vdash} $, defeated sequents may be derived from undefeated ones by any rule that expands the antecedent or defeater set of the conclusion. These include not only $\mathsf{LW},\mathsf{DE}$ but also the multiplicative binary rules ${\vee}{\vdash}, {\vdash}{\wedge},  {\neg}{\wedge}{\vdash}, {\vdash}{\neg}{\vee} $.

\begin{fact}[Derivation of Defeated Sequents]\label{fact:DeriDefeat}
	Defeated sequents can only be derived from (one or more) undefeated sequents via $\mathsf{LW},\mathsf{DE}, {\vee}{\vdash}, {\vdash}{\wedge},\\ {\neg}{\wedge}{\vdash}, {\vdash}{\neg}{\vee} $. 
\end{fact}

The engine of defeat in the calculus are the axioms and their respective defeater sets. The rules $ ax_1, ax_2 $ and $ ax_3 $ provide sequents with equivalent, possibly empty defeater sets. The members of these sets encode (extra-logical) conditions under which an agent is not entitled to believe that the atom is true.  We restrict these sets to literal formulas and insist that they not include the atom that they introduce or its negation, i.e. $ p, \neg p\not\in\bigcup\mathbf{S} $. This constraint ensures that the axioms will not introduce defeated sequents. Moreover, permitting the negations of atoms in the defeater sets would be redundant; inconsistent antecedents uniformly yield defeated sequents, since $ A, \neg A \vdash_\mathsf{SC} X $ for any set $ X \in \mathcal{L}_d $. Among other things, this means that there is no provable \textit{defeasible} sequent in $ \mathsf{SC}^\mathbf{S} $ corresponding to the law of \textit{ex contradictione quod libet} and to this extent, the calculus behaves paraconsistently. 

However, initial sequents of the form $ p,\neg p\vdash $ derived by the rule $ ax_4 $ are exempted from this behavior and their defeater sets are empty. The reasons for this are both conceptual and technical. The technical reason is that, as just noted, any sequent with an inconsistent antecedent and nonempty defeater set is automatically defeated. The conceptual reason is that defeaters follow the conclusions of inferences, in the sense that they tell agents when they are not authorized to infer beliefs. So it is only when formulas occur on the right-hand side of the turnstile that defeaters specific to those formulas are relevant. The calculus' ``symmetric'' logical rules ensure that these defeaters travel with their respective formulas, so to speak, throughout a derivation. 

\begin{definition}[Axiom Defeater Sets]\label{def:const-axioms}
	For any atom $ p $, we assign a set of (sets of) defeaters, denoted $ \mathcal{S}_p $, such that $ \bigcup\mathcal{S}_p \subset (\mathcal{A}\cup \{\neg q \mid q\in \mathcal{A}\}) \backslash \{p, \neg p\} $.
\end{definition}

Given the ``symmetric'' structure of $\mathsf{SC}^\mathbf{S} $, this constraint guarantees that the end-sequent of any $ \mathsf{RW} $-free derivation will contain, in its defeater set, the defeaters of every atom that occurs in the succedent.

\begin{fact}\label{fact:atomdefeatpres}
	If $ \Gamma\sststile{\mathbf{S}}{}\Delta $ is the end-sequent of a $ \mathsf{RW} $-free derivation in $\mathsf{SC}^\mathbf{S} $, then $ \mathcal{S}_p \subset\mathbf{S}$ for all $ p\in \mathcal{A}(\Delta) $, where $ \mathcal{A}(\Delta) $ denotes the set of atoms that occur in any member of $ \Delta $.
\end{fact}

We close this section by establishing the soundness and completeness of the propositional calculus $\mathsf{SC}^\mathbf{S}$.

\begin{definition}[Compatibility, $ \succsim $ for $\mathsf{SC}^\mathbf{S}$]\label{def:compatinSCs}
	When a set of d-wffs, $X$, and a set of sets of d-wffs, $ \mathbf{S} $, fail to meet the conditions of defeat, they are said to be \textit{compatible}. We use the symbol `$ \succsim $' to denote this relationship, which we define formally as follows:
	$$ X\succsim\mathbf{S} \:\textit{iff}\text{\:\:there is }no\:\,\varUpsilon\in\mathbf{S}\text{\, such that \,} \Gamma\sststile{\mathbf{T}}{} \varUpsilon\:\text{is derivable in}\: \mathsf{SC}^\mathbf{S}.$$
	We omit set notation on the left-hand side when there is no threat of misunderstanding, writing $ X, A \succsim \mathbf{S\cup T} $ for $ X\cup\{A\} \succsim \mathbf{S}\cup\mathbf{T}  $.
	When this relation fails to hold, we say that $ X $ and $ \mathbf{S} $ are \textit{incompatible} and express this by writing: $ X\not\succsim\mathbf{S} $.
\end{definition}

Although derivable sequents correspond to classical entailments, not all derivations are proofs. Thus, to establish the completeness of the calculus, we must show that every derivable \textit{undefeated} sequent has a proof.

\begin{lemma}[Derivable undefeated sequents are provable]\label{lem:Derivable_undefeated_provable_SCs}
	If $ X\sststile{\mathbf{S}}{}\varUpsilon$ is undefeated (i.e. $ X\succsim\mathbf{S} $) and derivable in $\mathsf{SC}^\mathbf{S} $, then there is a proof of $ X\sststile{\mathbf{S}}{}\varUpsilon$ in $\mathsf{SC}^\mathbf{S} $.
	
	\begin{proof}
		We proceed by induction on proof-height. For the base case, $ X\sststile{\mathbf{S}}{}\varUpsilon$ is obtained by an application of an axiom, and since it is undefeated (in fact following the constraints imposed on axiom defeater sets, it must be undefeated), the result is a proof. For induction, assume that any undefeated sequent whose derivation height is $ \leq h$ has a proof and that $ X\sststile{\mathbf{S}}{}\varUpsilon$ has a derivation height of $ h+1 $. The proof for the inductive case utilizes the following technique. Take the derivation of $ X\sststile{\mathbf{S}}{}\varUpsilon$ and for every sequent that occurs in the proof, replace its defeater set with the empty set. Next, remove every instance of the  $ \mathsf{DE} $-rule---these are now redundant given that all sequents have empty defeater sets. Finally apply $ \mathsf{DE} $ to the end-sequent as needed to obtain its original defeater set. By hypothesis, the end-sequent is undefeated and all other sequents in the derivation are undefeated either because they are the result of applying $ \mathsf{DE} $ to obtain the original end-sequent or because their defeater sets are empty. The derivation thus obtained is a proof.
		
		We illustrate this technique with the following example.
		
		$$\begin{prooftree}[template = \small$\inserttext$]
			\Hypo{}
			\Infer1[$ ax. $]{p \sststile{\{\{t\}\}}{} p}
			\Hypo{}
			\Infer1[$ ax. $]{r \sststile{\{\{p\}\}}{} r}
			\Infer2[${\vdash}{\wedge}$]{ p, r \sststile{\{\{t\},\{p\}\}}{} p\wedge r}
			\Hypo{}
			\Infer1[$ ax. $]{q \sststile{\{\{s\}\}}{} q}
			\Infer1[$\mathsf{DE}$]{q \sststile{\{\{s\},\{q\}\}}{} q}
			\Infer2[${\vee}{\vdash}$]{ p\vee q, r \sststile{\{\{t\},\{p\},\{s\},\{q\}\}}{} p\wedge r, q}
		\end{prooftree}$$
		\vspace{2mm}
		
		\noindent Since $ p, r \sststile{\mathbf{S}}{} p$ is derivable in $ \mathsf{SC}^\mathbf{S} $  we know that $ p, r\not\succsim \{\{p\},\{s\}\} $ and that $ p, r \sststile{\{\{p\},\{s\}\}}{} p\wedge r $ is thus defeated. Similarly, $ q \sststile{\{\{s\},\{q\}\}}{} q $ is defeated. We now apply the technique described above to obtain a proof.
		
				$$\begin{prooftree}[template = \small$\inserttext$]
			\Hypo{}
			\Infer1[$ ax. $]{p \sststile{\emptyset}{} p}
			\Hypo{}
			\Infer1[$ ax. $]{r \sststile{\emptyset}{} r}
			\Infer2[${\vdash}{\wedge}$]{ p, r \sststile{\emptyset}{} p\wedge r}
			\Hypo{}
			\Infer1[$ ax. $]{q \sststile{\emptyset}{} q}
			\Infer2[${\vee}{\vdash}$]{ p\vee q, r \sststile{\emptyset}{} p\wedge r, q}
			\Infer1[$\mathsf{DE}$]{ p\vee q, r \sststile{\{\{t\}}{} p\wedge r, q}
			\Ellipsis{}{}
			\Infer1[$\mathsf{DE}$]{ p\vee q, r \sststile{\{\{t\},\{p\},\{s\},\{q\}\}}{} p\wedge r, q}
		\end{prooftree}$$
		\vspace{2mm}
	\end{proof}

\end{lemma}

\begin{lemma}[Soundness and Completeness of $\mathsf{SC}^\mathbf{S} $]\label{lem:Sound&CompleteSCs}
	$ X\sststile{\mathbf{S}}{}\varUpsilon$ is provable in $\mathsf{SC}^\mathbf{S} $ iff $ X\vDash\bigvee\varUpsilon $ and $ X\nvDash\bigvee(\bigcup\mathbf{S}) $.
	\begin{proof}
		($ \Rightarrow $) By Lemma \ref{lem:SCS=CPL} and Definition \ref{def:compatinSCs},  it follows that $ X\nvDash\bigvee(\bigcup\mathbf{S}) $ iff $X\succsim\mathbf{S}$. Since, according to Definition \ref{def:proof}, every provable sequent is derivable and undefeated (i.e. $X\succsim\mathbf{S}$), we obtain the desired  result via Lemmas \ref{lem:SCS=CPL} and \ref{lem:Derivable_undefeated_provable_SCs}.\\
		($ \Leftarrow $) From Lemmas \ref{lem:SCS=CPL} and \ref{lem:Derivable_undefeated_provable_SCs}.
	\end{proof}
\end{lemma}


\subsection{Erotetic Defeat in $\mathsf{SC}^{?}$}\label{subsec:ED}
We now extend $ \mathsf{SC}^\mathbf{S}$ to $\mathsf{SC}^{?}$, whose rules are presented in Figure \ref{fig:SC?}. The axioms for this extension remain unchanged, as do the logical rules that apply to d-wffs, now explicitly restricted to the latter. The rule for defeater set expansion, $ \mathsf{DE} $ is also preserved, but $ \mathsf{LW} $  and $\mathsf{RW} $ are generalized to apply to any formula in $ \mathcal{L} $, as ranged over by the variables $ F, G, H $. Finally, $\mathsf{SC}^{?}$ contains four rules that yield e-wffs in sequents. We discuss each in detail below, but first we define defeat in $\mathsf{SC}^{?}$.

We extend the notion of defeat conservatively to ensure that sequents which are defeated in $\mathsf{SC}^\mathbf{S}$ remain defeated in $\mathsf{SC}^{?}$. To that end, we introduce the function, $\mathcal{E}(\cdot)$, which maps subsets of $ \mathcal{L} $ to subsets of $\mathcal{L}_d $. This function allows us to define defeat in $\mathsf{SC}^{?}$ in terms of defeat in $\mathsf{SC}^\mathbf{S}$.

\begin{definition}[$\mathcal{E}(\cdot)$]\label{def:EClo}
	Recall that d-wffs are ranged over by the variables $ A,B,C $ (with subscripts). Let $ \mathcal{E}(\Gamma) $ denote the set that results from removing all e-wffs from $ \Gamma $ and replacing them with a disjunction of their declarative constituents, i.e.
	$$  \mathcal{E}(\Gamma) = \{A: A\in\Gamma\}\cup\{A_1{\vee}\!\mydots\!{\vee} A_n: \:?\{A_1,\ldots,A_n\} \in \Gamma\}$$
	
	\noindent We say that $ \mathcal{E}(\Gamma) $ \textit{declarativizes} $ \Gamma $.
\end{definition}

\begin{definition}[Defeat in $\mathsf{SC}^{?}$]\label{def:defeat} A sequent in $\mathsf{SC}^{?}$ is defeated just in case a sequent with the declarativized antecedent and a succedent composed of at least one member of the defeater set is derivable in $\mathsf{SC}^\mathbf{S} $.		
	$$\Gamma\sststile{\mathbf{S}}{}\Delta\text{\,\,is defeated \textit{iff}\,} \exists X\in\mathbf{S} \text{\,\,such that\,\,}  \mathcal{E}(\Gamma)\sststile{\mathbf{T}}{} X \text{\,is derivable in\,\,}\mathsf{SC}^\mathbf{S}.$$

\end{definition}

We are now able to see how erotetic defeat is represented in our calculus and to explain why defeater sets are drawn from subsets of $ \mathcal{P}(\mathcal{P}(\mathcal{L}_d)) $. Recall that, in general, an erotetic inference is defeated when an agent is licensed to believe a direct answer to the relevant question---i.e. the question to which the agent would otherwise be licensed to inquire into. In order to capture this phenomenon, we must represent situations in which an agent is entitled to inquire into a question on the grounds of her entitlement to believe that at least one of its answers is true but in which, if the agent \textit{were} to gain entitlement to a particular answer, she would lose her license to inquire into it. Erotetic inferences are permissible when an agent knows that at least one answer to its question is true without knowing which one, but are not permissible when the agent knows a particular answer. Representing information about defeaters as sets of sets allows us to capture these situations.	

Consider the following:
\begin{example}\label{ex:Q_in_succ}
	$  p\vee\neg p\sststile{\{\{p\}\{\neg p\}\}}{} \:?\{p, \neg p\} $
\end{example}

\noindent If the sequent in Example \ref{ex:Q_in_succ} is provable, then the agent's entitlement to $ p\vee\neg p $ licenses it to inquire into the question $ ?\{p, \neg p\} $, since it is entitled to believe that at least one of its answers is true. But were the agent to obtain information that licensed the belief that $ p $ or that $ \neg p $---represented by adding either d-wff to the antecedent---it would forfeit that entitlement. Placing each direct answer to the succedent-question as a singleton in the defeater set has the effect of maintaining entitlement to believe that at least one answer to a question is true while barring such entitlement when a particular direct answer is obtained. As we shall demonstrate, Example \ref{ex:Q_in_succ} is interpretable as an inference rule underwritten by the evocation relation.

Sequent-analogs to e$_{sr}$-implication are also available. The sequent in Example \ref{ex:Q_in_eimp} would be defeated were either $ q $ or $r $ (i.e. the direct answers to the implying question) added to its antecedent. 

\begin{example}\label{ex:Q_in_eimp}
	$  \neg p \vee q, p \vee r, ?\{q, r\}\sststile{\{\{q\}\{r\}\}}{} \:?\{p, \neg p\} $
\end{example}

%
%
%

\subsection{Rules for E-wffs in $\mathsf{SC}^{?}$}\label{subsec:SCp?rules}
Let us now turn to the e-wff rules in the calculus.	As set-theoretic notation often becomes cumbersome, we avail ourselves of the following abbreviations.
\begin{definition}[Abbreviations]\label{def:abbrev}
	\hspace{1mm}
	\begin{itemize}
		\item $ ?[A_{\mid n}] $ abbreviates $ ?\{A_{1},\ldots,A_{n}\} $.
		\item $ [A_{\mid n}] $ abbreviates $ \{A_{1},\ldots,A_{n}\} $.
		\item $ \Gamma_{\mid n} $ and $ \mathbf{S}_{\mid n} $ abbreviate $ \Gamma_1 \cup \ldots \cup \Gamma_n$ and $ \mathbf{S}_1 \cup \ldots \cup \mathbf{S}_n$, respectively.
		\item $ [\mathbb{A}_{\mid n}] $ abbreviates $ \{\{A_{1}\},\ldots,\{A_{n}\}\} $.
	\end{itemize}	
\end{definition}

$\mathsf{SC}^{?}$ provides right and left rules for e-wffs (i.e., $ {\vdash}{?_1} $,$ {\vdash}{?_2} $, $ {?}{\vdash_1} $, and $ {?}{\vdash_2}$). Since right and left rules for connectives in a sequent calculus correspond, respectively, to the introduction and elimination rules in natural deduction, we will say that the rules  $ {\vdash}{?_1} $ and $ {\vdash}{?_2} $ govern the introduction of e-wffs and that $ {?}{\vdash_1} $ and $ {?}{\vdash_2}$ govern their elimination. Two pairs of introduction and elimination rules are needed to capture the distinction between evocation and e-implication. Both inference types have questions as their conclusions, but while the premises of the latter also include a question, those of the former do not. To represent evocation, we need a rule that permits the introduction (resp. elimination) of e-wffs in sequents whose antecedents (resp. succedents) consist solely of d-wffs, while representing implication requires a rule that permits such introduction (resp. elimination) in cases when the antecedents (resp. succedents) also contain e-wffs. It is natural to think that beliefs alone license inquiry in a manner that differs from the way beliefs \textit{plus} prior inquisitive commitments, represented by questions, license further inquiry. Formulating two versions of each introduction and elimination rule is an elegant way to respect this difference.

Let's examine $ {\vdash}{?_1} $. 
$$\begin{prooftree}
\Hypo{ X\sststile{\mathbf{S} }{} A_1,\ldots,A_n, \Delta}
\Infer1[$\vdash?_{\scriptstyle 1}$]{X\sststile{\mathbf{S}\,\cup\,\{\{A_1\},\ldots,\{A_n\}\}}{}
	\: ?\{A_1,\ldots,A_n\}, \Delta}
\end{prooftree}$$
\vspace{.5mm}

\noindent Ignoring side formulas, we read the rule as saying that an agent entitled to believe that some statement in the set $ \{A_{1},\mydots,A_{n}\} $ is true is thereby entitled to inquire into the question $ ?\{A_{1},\mydots,A_{n}\} $, so long as she is not entitled to believe any particular member of that set, i.e. any direct answer to the question.

While this rule is intended to capture evocation, it actually represents a more general relationship, since the succedents of the premise and conclusion may include a set of side formulas composed of d-wffs or e-wffs. The generality of the rule is motivated by the desire to formulate rules for e-wffs that conform to the standard format of multiple-succedent calculi.

When there are no side formulas, the rule $\vdash?_{\scriptstyle 1}$ faithfully depicts evocation, as demonstrated by Theorem \ref{thrm:S+C_EE} below. The first condition in Definition \ref{def:erotetic_evocation} is satisfied by the fact that the succedent of the premise, i.e.$  \{A_1,\mydots,A_n\} $,  contains all of the direct answers to the question in the succedent of the conclusion, i.e.,  $?\{A_1,\mydots,A_n\} $. Satisfying the second condition of erotetic evocation, i.e. the one concerning its defeasibility, requires significant exploitation of the defeater set mechanism. Recall that this condition prohibits evocation whenever the set of d-wffs entails a direct answer to the evoked question. By including singletons of each direct answer in the defeater set of the conclusion, $ \vdash ?_1 $ guarantees that the set of auxiliary d-wffs does not entail a direct answer to the evoked question. Since defeat is defined as a relation between the d-wffs in the antecedent set on one hand, and members of the defeater set on the other, the inclusion of direct answers as separate sets permits $ X $ to fulfill the second condition without jeopardizing its satisfaction of the first.

Let us now consider the left-rule for e-wffs, $ {?}{\vdash_1} $. 
$$\begin{prooftree}
\Hypo{\Gamma_1, A_1 \sststile{\mathbf{S}_1}{} X_1}
\Hypo{\hspace{-5.5mm}\ldots\Gamma_n, A_n \sststile{\mathbf{S}_n }{} X_n}
\Infer2[${?}{\vdash_{\scriptstyle 1}}$]{ \Gamma_{\mid n}, ?\{A_1,\ldots,A_n\} \sststile{\mathbf{S}_{\mid n}\,\cup\,\{\{A_1\},\ldots,\{A_n\}\}}{}X_{\mid n}}
\end{prooftree}$$

Again ignoring side formulas, the rule says that an agent entitled to believe at least one member of a set of statements, $ X $, on the basis of its entitlement to believe $ A_1 $ or \ldots or $ A_n $ is also entitled to $ X $ on the basis of its entitlement to inquire into the question $ ?\{A_{1},\mydots,A_{n}\} $, so long as it is not licensed to believe one of its answers.

The observant reader will no doubt have noticed that $ {\vdash}{?_1} $ and $ {?}{\vdash_1} $ are essentially expanded versions of the classical elimination rules for $ \vee $ in $ \mathsf{SC}^\mathbf{S} $. This overlap is to be expected, since questions in IEL for classical logic \textit{presuppose} that at least one of their possible answers is true. However, the addition of formulas to the conclusion's defeater set shows that $ {\vdash}{?_1} $ (resp. $ {?}{\vdash_1} $) is not equivalent to $ \vdash\vee $ (resp. $ \vee\vdash $).

We now turn to the second introduction rule for e-wffs, i.e., $ \vdash ?_2 $. 
$$\begin{prooftree}[template = \small$\inserttext$]
\Hypo{\Gamma_1, ?[A_{\mid n}]\sststile{\mathbf{S}}{}[B_{\mid m}], \Delta_1}
\Hypo{\Gamma_2, B_{1}\sststile{\mathbf{T}_{1} }{} A_i, \Delta_2}
\Hypo{\hspace{-6.5mm}\ldots\Gamma_{m+1}, B_{m}\sststile{\mathbf{T}_{m} }{} A_{j}, \Delta_{m+1}}
\Infer3[${\vdash}{?_{\scriptstyle 2}}$]{ \Gamma_{\mid m+1} , ?[A_{\mid n}] \sststile{\mathbf{S}\,\cup\,\mathbf{T}_{\mid m}\cup\,\{\{A_1\},\ldots,\{A_n\}\}}{}  \,?[B_{\mid m}], \Delta_{\mid m+1}}
\end{prooftree}$$

This rule may be read  top-down, omitting side formulas,  as follows: anyone entitled to believe that $ B_1 $ or \ldots or $ B_m $ on the basis of its entitlement to inquire into $ ?\{A_{1},\mydots,A_{n}\} $ and whose entitlement to believe any member of $ \{B_{1},\mydots,B_{n}\} $ licenses an answer to that question (i.e. $ \{A_i,\mydots,A_j\} \subseteq \{A_{1},\mydots,A_{n}\}$), is thereby entitled to inquire into $  ?\{B_{1},\mydots,B_{n}\}$, so long as she is not licensed to believe one of its answers. The rule is intended to capture the relation of e$_{sr}$-implication. The left-most premise roughly corresponds to the first clause in Definition \ref{def:sr_e-imp}, according to which an answer to the implying question, together with the set of d-wffs, entails that there is a true answer to the implied question. The premises to the right roughly correspond to the second condition, according to which, an answer to the implied question, together with the set of d-wffs, entails an answer to the implying question. The presence of the direct answers to $ ?\{A_{1},\mydots,A_{n}\} $ as singletons in the conclusion's defeater set attempts to capture e$_{sr}$-implication as stated in Definition \ref{def:sr_e-imp}'s third clause. We note that the d-wffs that appear in the succedent of the second set of premise, i.e. $ A_i,\mydots,A_j $, may include any of those that occur in $ \{A_{1},\mydots, A_{n}\} $. They need not exhaust the latter and it is perfectly acceptable that $ i=j $. This applies to the proviso on $ {?}{\vdash_2} $ as well. 

Admittedly, $ {\vdash}{?_2} $ encodes less restrictive conditions than those imposed by e$_{sr}$-implication insofar as it permits side formulas of either the declarative or erotetic sort. Again, this liberalization brings the rule into conformity with standard rules in classical sequent calculi.


Finally, we turn to $ {?}{\vdash_2} $ (see Figure \ref{fig:SC?}). We read the rule top-down, omitting side formulas, as follows: an agent who is entitled to inquire into $ ?\{B_1,\mydots, B_m\} $ on the basis of entitlement to believe any statement from the set $\{A_{1},\mydots, A_{n}\} $ and who is entitled to believe some statement in the latter on the basis of her entitlement to believe any statement in $ \{B_1,\mydots, B_m\} $, is thereby licensed to inquire into $ ?\{B_1,\mydots, B_m\} $ if she is entitled to inquire into $ ?\{A_{1},\mydots, A_{n}\} $, so long as she is not licensed to believe one of its answers. Recall that the motivation behind $ \vdash ?_1 $ and $ \vdash ?_2 $ is that in order to represent evocation and e$_{sr}$-implication, a calculus needs to distinguish between inferences in which a question follows from a set of statements and those in which a question follows from at least one question. Symmetry demands that we respect this difference in our elimination rules as well.

\section{Main Results}\label{sec:Results}

We move now to proving that certain classes of sequents in $\mathsf{SC}^{?}$ are sound and complete with respect to evocation and e$_{sr}$-implication. We begin with some helpful Lemmas.

\begin{definition}[Compatibility, $ \succsim $, in $ \mathsf{SC}^? $]\label{def:compat_in_SC?}
When a set of formulas, $ \Gamma $, and a set of sets of d-wffs, $ \mathbf{S} $, fail to meet the conditions of defeat, they are said to be \textit{compatible}. We use the symbol `$ \succsim $' to denote this relationship, which we define formally as follows:
$$ \Gamma\succsim\mathbf{S} \:\textit{iff}\text{\:\:there is no}\:\,X\in\mathbf{S}\text{\, such that \,} \mathcal{E}(\Gamma)\sststile{\mathbf{T}}{} X\:\text{is derivable in}\: \mathsf{SC}^\mathbf{S}.$$
Again, we use the comma for set union when there is no threat of misunderstanding.
\end{definition}

\begin{definition}[Provability of sequents in $ \mathsf{SC}^{?} $]
We write $\Gamma \sststile{\mathbf{S}}{\mathsf{SC}^{?}} \Delta $ to denote the fact that there is a proof of $ \Gamma \sststile{\mathbf{S}}{} \Delta $ in $ \mathsf{SC}^{?} $.
\end{definition}


\begin{lemma}[Height-Preserving Invertibility of $ {\vee}{\vdash} $]\label{lem:invert-LV}
If $ \Gamma, A_1{\vee}\!\mydots\!{\vee} A_n\sststile{\mathbf{S}}{}  \Delta $ has a proof height $ \leq h $, then $ \Gamma', A_i \sststile{\mathbf{S'}}{} \Delta' $ for all $ i\in\{1,\mydots,n\} $ and some $ \Gamma'\subseteq \Gamma, \Delta' \subseteq\Delta, \mathbf{S'}\subseteq\mathbf{S}  $ have proofs of height $ \leq h $.

\begin{proof}
	By induction on proof height. For the base case, $ \Gamma, A_1{\vee}\!\mydots\!{\vee} A_n\sststile{\mathbf{S}}{} \Delta  $ cannot be obtained directly from an axiom.  For induction, assume that height-preserving inversion holds up to height $  h $ and that $ \Gamma, A_1{\vee}\!\mydots\!{\vee} A_n\sststile{\mathbf{S}}{} \Delta $ has a proof of height $ h+1 $. From Definition \ref{def:proof}, it follows by this hypothesis that the premises employed to derive $ \Gamma, A_1{\vee}\!\mydots\!{\vee} A_n\sststile{\mathbf{S}}{} \Delta $ are undefeated, and thus we need only show that there is a height-preserving derivation. There are two cases:\\
	\textit{Case 1:} $ A_1{\vee}\!\mydots\!{\vee} A_n $ is principal in the last rule. It must be obtained via $ \mathsf{LW} $ or $ {\vee}{\vdash} $. Assume the latter. So, $ \Gamma_1, A_1 \sststile{\mathbf{S}_1}{} \Delta_1 $ and $\Gamma_n, A_2{\vee}\!\mydots\!{\vee} A_n \sststile{\mathbf{S}_n }{}\Delta_n $, where $ \mathbf{S}_{\mid n} = \mathbf{S}; \Gamma_{\mid n} = \Gamma;$ and $ \Delta_{\mid n} = \Delta$, have derivations $ \leq h$. The first of these sequents satisfies the claim's consequent, while the second is covered by our inductive hypothesis.  If $ A_1{\vee}\!\mydots\!{\vee} A_n $ is obtained by $ \mathsf{LW} $, then $ \Gamma \sststile{\mathbf{S}}{} \Delta  $ has a derivation of height $ \leq h $. We apply $ \mathsf{LW} $ to obtain the sequents $ \Gamma, A_i \sststile{\mathbf{S}}{} \Delta $ for all $ i\in\{1,\mydots,n\} $ with proof heights $ \leq h+1 $.
	\textit{Case 2:} $ A_1{\vee}\!\mydots\!{\vee} A_n$ is not principal in the last rule. Then it has no more than two premises $ \Gamma'',  A_1{\vee}\!\mydots\!{\vee} A_n\sststile{\mathbf{S''}}{} \Delta'' $ and $ \Gamma''',  A_1{\vee}\!\mydots\!{\vee} A_n\sststile{\mathbf{S'''}}{} \Delta''' $ with proof heights $ \leq h $, so by inductive hypothesis  $ \Gamma'', A_i \sststile{\mathbf{S''}}{} \Delta''  $ and $ \Gamma''', A_i \sststile{\mathbf{S'''}}{} \Delta''' $ for all $ i\in\{1,\mydots,n\} $ have proofs of height $ \leq h $.
\end{proof}
\end{lemma}

\begin{lemma}[Height-Preserving Invertibility of $ {\vdash}{\vee} $]\label{lem:invert-RV}
 If $ \Gamma\sststile{\mathbf{S}}{}\! A_1{\vee}\!\mydots\!{\vee} A_n, \Delta $ has a proof height $ \leq h $, then $ \Gamma\sststile{\mathbf{S}}{} A_1,\mydots, A_n, \Delta $ has a proof of height $ \leq h $.

\begin{proof}
	By induction on proof height. Assume height-preserving inversion up to $ h $ and let $ \Gamma\sststile{\mathbf{S}}{} A_1{\vee}\!\mydots\!{\vee} A_n, \Delta $ have a proof height $ h+1 $. Again, it follows by this hypothesis that the premises used to derive this sequent are undefeated. If $ A_1\vee\mydots\vee A_n $ is the principal formula in the last rule, then $\Gamma\sststile{\mathbf{S}}{} A_1,\mydots, A_n, \Delta$ has a proof of $ \leq h $. If $ A_1\vee\mydots\vee A_n $ is not principal in the last rule, then the conclusion follows from no more than two premises $\Gamma''\sststile{\mathbf{S''}}{} A_1\vee\mydots\vee A_n, \Delta'' $ and $\Gamma'''\sststile{\mathbf{S'''}}{} A_1\vee\mydots\vee A_n, \Delta''' $ with proof heights $ \leq h $. By inductive hypothesis, it follows that $ \Gamma''\sststile{\mathbf{S''}}{} A_1,\mydots, A_n, \Delta'' $ and $ \Gamma'''\sststile{\mathbf{S'''}}{} A_1,\mydots, A_n, \Delta''' $ have proofs of height $ \leq h $. Apply the last rule to obtain $ \Gamma\sststile{\mathbf{S}}{} A_1,\mydots, A_n, \Delta $ with a proof of height $ \leq h+1 $.
\end{proof}
\end{lemma}

\begin{lemma}\label{lem:lero}
$ \Gamma, ?[A_{\mid n}]\sststile{\mathbf{S}}{\mathsf{SC}^{?}} \varUpsilon \:\: $ iff $ \:\: \Gamma, A_1{\vee}\!\mydots\!{\vee} A_n\sststile{\mathbf{S}}{\mathsf{SC}^{?}} \varUpsilon $.
\begin{proof}
	($ \Rightarrow $) By induction on proof height. For the base case, $ \Gamma, ?[A_{\mid n}]\sststile{\mathbf{S}}{} \varUpsilon  $ is not an axiom as it contains at least one e-wff. For induction, assume our result holds for all proofs of height $ \leq h $ and that  $ \Gamma, ?[A_{\mid n}]\sststile{\mathbf{S}}{} \varUpsilon $ has a proof of height $ h+1 $. There are two cases.\\
	\textit{Case 1:} If $ ?[A_{\mid n}] $ is the principal formula, then it is obtained via $ \mathsf{LW} $ or $ {?}{\vdash_1} $, since the succedent is declarative. Assume it is obtained by the latter. It follows that the premises of $ {?}{\vdash_1} $ are provable, i.e. $ \Gamma_1, A_1 \sststile{\mathbf{T}_1}{\mathsf{SC}^{?}} \varUpsilon_1 \ldots \Gamma_n, A_n \sststile{\mathbf{T}_n }{\mathsf{SC}^{?}} \varUpsilon_n $, where $ \Gamma_{\mid n} = \Gamma;$ $ \varUpsilon_{\mid n} = \varUpsilon$ and $ \mathbf{T}_{\mid n}\cup[\mathbb{A}_{\mid n}] = \mathbf{S}$. Since none of these premises is defeated, there is no $ i\in\{1,\mydots,n\} $ such that $A_i\in\bigcup\mathbf{T}_i $. Successive applications of $ {\vee}{\vdash} $ and appropriate applications of $ \mathsf{DE} $ to these sequents (e.g. to add $ [\mathbb{A}_{\mid n}] $ to the defeater set if it is not already included) yield $ \Gamma, A_1{\vee}\!\mydots\!{\vee} A_n\sststile{\mathbf{S}}{} \varUpsilon $. Since, by hypothesis, $ \Gamma, ?[A_{\mid n}]\sststile{\mathbf{S}}{} \varUpsilon $ is undefeated and since it follows from Definition \ref{def:EClo} that $\mathcal{E}(\Gamma\,\cup\,?[A_{\mid n}])= \mathcal{E}(\Gamma)\,\cup\, A_1{\vee}\!\mydots\!{\vee} A_n $, we know that $ \Gamma\,\cup\, \{A_1{\vee}\!\mydots\!{\vee} A_n\}\succsim \mathbf{S}$. So, $ \Gamma, A_1{\vee}\!\mydots\!{\vee} A_n\sststile{\mathbf{S}}{\mathsf{SC}^{?}} \varUpsilon $. If $ ?[A_{\mid n}] $ is introduced via $ \mathsf{LW} $, then $ \Gamma \sststile{\mathbf{S}}{\mathsf{SC}^{?}} \varUpsilon $. We apply $ \mathsf{LW} $  to obtain a proof of $ \Gamma, A_1{\vee}\!\mydots\!{\vee} A_n\sststile{\mathbf{S}}{} \varUpsilon $. \\
	\textit{Case 2:} If $ ?[A_{\mid n}] $ is not principal in the last rule, then it follows from no more than two premises $ \Gamma', ?[A_{\mid n}]\sststile{\mathbf{S'}}{} \varUpsilon' $ and $ \Gamma'', ?[A_{\mid n}]\sststile{\mathbf{S''}}{} \varUpsilon'' $ with proof heights $ \leq h $. So by inductive hypothesis, $ \Gamma',  A_1{\vee}\!\mydots\!{\vee} A_n\sststile{\mathbf{S'}}{\mathsf{SC}^{?}} \varUpsilon' $ and $ \Gamma'',  A_1{\vee}\!\mydots\!{\vee} A_n\sststile{\mathbf{S''}}{\mathsf{SC}^{?}} \varUpsilon'' $. We apply the last rule to obtain a proof of $ \Gamma, A_1{\vee}\!\mydots\!{\vee} A_n\sststile{\mathbf{S}}{} \varUpsilon $.
	
	($ \Leftarrow $) Again, by induction on proof height. From Lemma \ref{lem:invert-LV}, we know that  $ \Gamma', A_i \sststile{\mathbf{S'}}{\mathsf{SC}^{?}} \varUpsilon' $ for all $ i\in\{1,\mydots,n\} $ and some $ \Gamma'\subseteq \Gamma, \varUpsilon' \subseteq\varUpsilon, \mathbf{S'}\subseteq\mathbf{S}$ have proof heights $ \leq h $. We apply $ {?}{\vdash_1} $ to these sequents to obtain a proof of $ \Gamma, ?[A_{\mid n}]\sststile{\mathbf{S}}{} \varUpsilon $.
\end{proof}
\end{lemma}

\begin{lemma}\label{lem:rero}
$ X\sststile{\mathbf{S}}{\mathsf{SC}^{?}} ?[B_{\mid m}], \Delta \:\: $ iff $ \:\: X\sststile{\mathbf{S}}{\mathsf{SC}^{?}} B_1{\vee}\!\mydots\!{\vee} B_m, \Delta $.
\begin{proof}
	($ \Rightarrow $) By induction on proof height. In the base case, $ X\sststile{\mathbf{S}}{} ?[B_{\mid m}], \Delta $ is not an axiom. For induction, we assume that the result holds for all proofs of height $ \leq h $ and that $ X\sststile{\mathbf{S}}{} ?[B_{\mid m}], \Delta $ has a proof of height $ h+1 $. Again there are two cases.\\
	\textit{Case 1:} If $ ?[B_{\mid m}] $ is the principal formula, then it must be obtained via $ \mathsf{RW} $ or $ {\vdash}{?_1} $, since the antecedent is declarative. If it follows via the latter, the premise is provable, so $ X\sststile{\mathbf{T}}{\mathsf{SC}^{?}} B_1,\mydots,B_m, \Delta $ where $ \mathbf{T}\cup[\mathbb{B}_{\mid m}]=\mathbf{S} $.  We apply $ {\vdash}{\vee} $, followed, if needed, by $ \mathsf{DE} $ to add $[\mathbb{B}_{\mid m}] $ to $ \mathbf{T} $, and thus obtain a proof of $ X\sststile{\mathbf{S}}{\mathsf{SC}^{?}} B_1{\vee}\!\mydots\!{\vee} B_m, \Delta $. If $ ?[B_{\mid m}] $ is introduced via $ \mathsf{RW} $, then $ X\sststile{\mathbf{S}}{\mathsf{SC}^{?}} \Delta $. Apply $ \mathsf{RW} $ to weaken the succedent by $ B_1{\vee}\!\mydots\!{\vee} B_m $ and we have the desired proof.\\
	\textit{Case 2:} If $ ?[B_{\mid m}] $ is the not principal in the last rule, then the conclusion follows from no more than two premises $X'\sststile{\mathbf{S'}}{} ?[B_{\mid m}], \Delta' $ and $X''\sststile{\mathbf{S''}}{} ?[B_{\mid m}], \Delta'' $ with proof heights $ \leq h $. So by inductive hypothesis $X'\sststile{\mathbf{S'}}{\mathsf{SC}^{?}} B_1{\vee}\!\mydots\!{\vee} B_m, \Delta' $ and $X''\sststile{\mathbf{S''}}{\mathsf{SC}^{?}} B_1{\vee}\!\mydots\!{\vee} B_m, \Delta'' $. We apply the last rule to obtain $ X\sststile{\mathbf{S}}{\mathsf{SC}^{?}} B_1{\vee}\!\mydots\!{\vee} B_m, \Delta $.
	
	($ \Leftarrow $) Again, by induction on proof height. From Lemma \ref{lem:invert-RV}, we know that $ \Gamma\sststile{\mathbf{S}}{\mathsf{SC}^{?}} B_1,\ldots, B_m, \Delta $ has a proof of height $ \leq h $.  We apply $ {\vdash}{?_1} $ to obtain the desired sequent.
\end{proof}
\end{lemma}

\begin{lemma}\label{lem:bero}
$ X, ?[A_{\mid n}]\sststile{\mathbf{S}}{\mathsf{SC}^{?}} ?[B_{\mid m}], \varUpsilon \:$ iff\\
$(i)\: X, A_1{\vee}\!\mydots\!{\vee} A_n\sststile{\mathbf{S}}{\mathsf{SC}^{?}} B_1{\vee}\!\mydots\!{\vee} B_m, \varUpsilon $; and\\ 
(ii)\: for all $ j\in\{1,\mydots,m\} $, there exists $ k\in\{1,\mydots,n\} $ and exists $\mathbf{T} \subset \mathbf{S}$ such that $X, B_j \sststile{\mathbf{T}}{\mathsf{SC}^{?}} A_k , \varUpsilon $.\\ 

\begin{proof}
	$ (\Rightarrow) $ By induction on proof height. There are two cases.\\
	\textit{Case 1:} Either $ ?[A_{\mid n}]$ or $ ?[B_{\mid m}]$ is principal in the last rule. There are three sub-cases.\\
	(1a.) Suppose that the last rule applied is  $ {?}{\vdash_2} $ and $ ?[A_{\mid n}]$ is principal. So, there are proofs of the left-hand set of premises in $ {?}{\vdash_2} $, i.e. sequents of the form $ X_i, A_i\sststile{\mathbf{S}_i}{} ?[B_{\mid m}], \varUpsilon_i $ for all $ i \in \{1,\ldots,n\} $. Iterative applications of $ {\vee}{\vdash} $ yield $ X', A_1{\vee}\!\mydots\!{\vee} A_n\sststile{\mathbf{S'}}{} ?[B_{\mid m}], \varUpsilon' $ where $ X'\subseteq X$, $\varUpsilon' \subseteq\varUpsilon$ and $ \mathbf{S'}\subseteq\mathbf{S}$. From the appropriate applications of $ \mathsf{LW}, \mathsf{RW} $ and $ \mathsf{DE} $ we obtain $ X, A_1{\vee}\!\mydots\!{\vee} A_n\sststile{\mathbf{S}}{} ?[B_{\mid m}], \varUpsilon$.  From Lemma \ref{lem:rero}, it follows that $ X, A_1{\vee}\!\mydots\!{\vee} A_n\sststile{\mathbf{S}}{\mathsf{SC}^{?}} B_1{\vee}\!\mydots\!{\vee} B_m, \varUpsilon$, and condition (i) is satisfied. Similarly, it follows that there are proofs of the right-hand set of premises in $ {?}{\vdash_2} $, i.e. for all $ j\in\{1,\mydots,m\}$, there exists $ k\in\{1,\mydots,n\} $, $ X'\subseteq X, \varUpsilon'\subseteq\varUpsilon, $ and $ \mathbf{T}_{j}\subset\mathbf{S}$ such that $ X',  B_j\sststile{\mathbf{T}_j}{} A_k, \varUpsilon'$. Note that since none of these premises is defeated, $ A_k\not\in\bigcup\mathbf{T}_j $. As $ \{A_k\}\in[\mathbb{A}_{\mid n}] $ and $[\mathbb{A}_{\mid n}]\subseteq\mathbf{S}$, the defeater sets of these sequents must be \textit{proper} subsets of the end-sequent's defeater set, i.e. for all $ j\in\{1,\mydots,m\}$, $\mathbf{T}_j\subset\mathbf{S}$. By judicious applications of $ \mathsf{LW}, \mathsf{RW}, $ and $ \mathsf{DE} $ we arrive at proofs of the sequents $ X,  B_j\sststile{\mathbf{T}}{} A_k, \varUpsilon$ for all $ j\in\{1,\mydots,m\}$ and at least one $ k\in\{1,\mydots,n\} $, where $ \mathbf{T} $---for reasons just mentioned---is some proper subset of $ \mathbf{S}$. Thus, condition (ii) is satisfied.\\
	(1b.) Suppose that the last rule applied is ${\vdash}{?_2} $ and $ ?[B_{\mid m}]$ is principal.  It follows that the left-hand sequent of that rule, $ X', ?[A_{\mid n}]\sststile{\mathbf{S}'}{} B_1,\mydots, B_m, \varUpsilon' $ is provable, where $ X'\subseteq X$, $\varUpsilon' \subseteq\varUpsilon$ and $ \mathbf{S'}\subseteq\mathbf{S} $. We apply $ {\vdash}{\vee} $ to obtain $ X', ?[A_{\mid n}]\sststile{\mathbf{S}'}{} B_1{\vee}\!\mydots\!{\vee} B_m, \varUpsilon' $ and from Lemma \ref{lem:lero} we have $ X',  A_1\vee\mydots\vee A_n\sststile{\mathbf{S}'}{} B_1{\vee}\!\mydots\!{\vee} B_m, \varUpsilon' $. Apply $ \mathsf{LW}, \mathsf{RW}, $ and $ \mathsf{DE} $ as needed to obtain $  X, A_1\vee\mydots\vee A_n\sststile{\mathbf{S}}{} B_1{\vee}\!\mydots\!{\vee} B_m, \varUpsilon $, which satisfies condition (i). The right-hand set of premises in ${\vdash}{?_2} $ ensure that condition (ii) is satisfied in the manner of the first sub-case.\\
	(1c.) If $ ?[A_{\mid n}]$ or $ ?[B_{\mid m}]$ is principle and the last rule is weakening, then the result follows by inductive hypothesis in the manner of Lemmas \ref{lem:lero} and \ref{lem:rero}.\\
	\textit{Case 2:} Neither $ ?[A_{\mid n}]$ nor $ ?[B_{\mid m}]$ is principal in the last rule. So $ X, ?[A_{\mid n}]\sststile{\mathbf{S}}{} ?[B_{\mid m}], \varUpsilon $ follows from one or more premises covered by the inductive hypothesis.
	
	($ \Leftarrow $) By induction on proof height of sequents in conditions (i) and (ii). Given condition (i), it follows from Lemma \ref{lem:invert-RV} that $ X, A_1{\vee}\!\mydots\!{\vee} A_n \sststile{\mathbf{S}}{\mathsf{SC}^{?}} B_1,\mydots, B_m,\Delta $ and from Lemma \ref{lem:lero} we obtain	
	$ X, ?[A_{\mid n}] \sststile{\mathbf{S}}{} B_1,\mydots, B_m,\Delta $. Condition (ii) then gives us the premises needed to apply $ {\vdash}{?_2} $, which yields $ X, ?[A_{\mid n}]\sststile{\mathbf{S}}{} ?[B_{\mid m}], \varUpsilon $.  			 
\end{proof}
\end{lemma}

\begin{corollary}\label{coro:declaredseq}
If $\Gamma\sststile{\mathbf{S}}{\mathsf{SC}^{?}}\Delta $,\: then\:\: $ \mathcal{E}(\Gamma)\sststile{\mathbf{S}}{\mathsf{SC}^{?}}\mathcal{E}(\Delta) $.

\begin{proof}			
	Let $ \mathsf{E}(\cdot) $ be a function $ \mathcal{P}(\mathcal{L})\mapsto \mathbb{N} $ that returns the number of e-wffs in a set of formulas. We proceed by induction on the number of e-wffs in the left-hand  sequent, i.e. $ \mathsf{E}(\Gamma\cup\Delta) $. For our base case, $  \mathsf{E}(\Gamma\cup\Delta) = 0$. So $ \Gamma\sststile{\mathbf{S}}{}\Delta $ contains no e-wffs and thus satisfies the conditional. For induction, assume our result holds for all sequents such that $ \mathsf{E}(\Gamma\cup\Delta) \leq n $. Now assume that $ \Gamma\sststile{\mathbf{S}}{}\Delta $ is such that $\mathsf{E}(\Gamma\cup\Delta) = n+1$. It follows from our inductive hypothesis that $ X\sststile{\mathbf{S}}{\mathsf{SC}^{?} }\varUpsilon  $  where  either $ X =   \mathcal{E}(\Gamma\backslash\{Q\})$ or $\varUpsilon =   \mathcal{E}(\Delta\backslash\{Q\})$ for some e-wff $ Q \in\Gamma\cup\Delta $. Thus we need only show that $ X\cup\mathcal{E}(\{Q\})\sststile{\mathbf{S}}{\mathsf{SC}^{?}}\varUpsilon  $ and $ X\sststile{\mathbf{S}}{\mathsf{SC}^{?}}\varUpsilon\cup\mathcal{E}(\{Q\})$ and these follow from Lemmas \ref{lem:lero} and \ref{lem:rero}, respectively.
\end{proof}
\end{corollary}

\begin{corollary}\label{coro:conversext}
$\mathsf{SC}^{?}$ is a conservative extension of $\mathsf{SC}^{\mathbf{S}}$, that is, any sequent composed solely of d-wffs (i.e. formulas of $ \mathcal{L}_d $) that is provable in $\mathsf{SC}^{?}$is provable in $\mathsf{SC}^{\mathbf{S}}$ and vice versa.
\end{corollary}
Corollary \ref{coro:declaredseq} tells us that the declarativized version of any provable erotetic sequent (i.e. a sequent containing at least one e-wff) is also provable. This is not particularly surprising given the tight relationship between classical disjunctions and e-wffs.	From the adequacy of the declarative calculus $ \mathsf{SC}^\mathbf{S} $ for CPL (i.e. Lemma \ref{lem:SCS=CPL}), it follows that two classes of defeasible sequents in  $ \mathsf{SC}^{?}$ are sound and complete with respect to evocation (Definition \ref{def:erotetic_evocation}) and strong regular e-implication (Definition \ref{def:sr_e-imp}).

\begin{theorem}\label{thrm:S+C_EE}
$  X \rightarrowtail ?[A_{\mid n}]\: $ iff $\:X\sststile{\mathbf{S}\cup [\mathbb{A}_{\mid n}]}{\mathsf{SC}^{?}} \: ?[A_{\mid n}] $ for some (possibly empty) $ \mathbf{S}$.
\begin{proof}
	($\Rightarrow$) From Lemma \ref{lem:Sound&CompleteSCs}, the first condition in Definition \ref{def:erotetic_evocation} implies that $ X\sststile{\mathbf{S}}{\mathsf{SC}^{?}} A_1 \vee\mydots\vee A_n $ for some (possibly empty) $ \mathbf{S}$ and the second condition implies that $ X\succsim [\mathbb{A}_{\mid n}]$. Apply $ \mathsf{DE} $ to  $ X\sststile{\mathbf{S}}{} A_1 \vee\mydots\vee A_n $ to obtain a proof of  $ X\sststile{\mathbf{S}\cup[\mathbb{A}_{\mid n}]}{} A_1 \vee\mydots\vee A_n $. The result then follows by Lemma \ref{lem:rero}.\\
	($ \Leftarrow $) From Lemma \ref{lem:rero}, it follows by hypothesis that $ X\sststile{\mathbf{S}\cup[\mathbb{A}_{\mid n}]}{\mathsf{SC}^{?}} A_1 \vee\mydots\vee A_n $. Thus, $ X \succsim [\mathbb{A}_{\mid n}]$. By Definitions \ref{def:defeat} and \ref{def:compat_in_SC?} and Lemma \ref{lem:Sound&CompleteSCs}, it follows that the two conditions on erotetic evocation (Definition \ref{def:erotetic_evocation}) are satisfied.
\end{proof}
\end{theorem}

\begin{theorem}\label{thrm:S+C_SR-EI}
$X\mid\: ?[ A_{\mid n}] \twoheadrightarrow \:?[ B_{\mid m}]$ \:iff\: $ X, ?[ A_{\mid n}]\sststile{\mathbf{S}\cup [\mathbb{A}_{\mid n}]}{\mathsf{SC}^{?}} \: ?[ B_{\mid m}]$ for some (possibly empty) $ \mathbf{S}$.
\begin{proof}
	($\Rightarrow$) By Definition \ref{def:sr_e-imp} and Lemma \ref{lem:Sound&CompleteSCs}, it follows that (i) $ X, A_1{\vee}\!\mydots\!{\vee} A_n\sststile{\mathbf{S}\cup[\mathbb{A}_{\mid n}]}{\mathsf{SC}^{?}} B_1\vee\mydots\vee B_m $, (since the third condition in Definition \ref{def:sr_e-imp} entails that $ X\succsim [\mathbb{A}_{\mid n}]$) and that (ii) for all $ j\in\{1,\mydots,m\} $, there exists $ k\in\{1,\mydots,n\} $ and exists $ \mathbf{T}\subset\mathbf{S}\cup[\mathbb{A}_{\mid n}] $ such that $ X, B_j \sststile{\mathbf{T}}{\mathsf{SC}^{?}} A_k  $. The result is then obtained by Lemma \ref{lem:bero}.\\			
	($ \Leftarrow $) From Lemma \ref{lem:bero}, it follows that (i) $ X, A_1{\vee}\!\mydots\!{\vee} A_n\sststile{\mathbf{S}\cup[\mathbb{A}_{\mid n}]}{\mathsf{SC}^{?}} B_1{\vee}\!\mydots\!{\vee} B_m $ and that (ii) for all $ j\in\{1,\mydots,m\} $, there exists $ k\in\{1,\mydots,n\} $ and exists $ \mathbf{T}\subset\mathbf{S}\cup[\mathbb{A}_{\mid n}] $ such that $ X, B_j \sststile{\mathbf{T}}{\mathsf{SC}^{?}} A_k  $. From Definition \ref{def:sr_e-imp} and Lemma \ref{lem:Sound&CompleteSCs} we know that the three conditions on e$_{sr}$-implication (Definition \ref{def:sr_e-imp}) are satisfied.
\end{proof}
\end{theorem}

\begin{theorem}\label{thrm:decidable}
$ \mathsf{SC}^?$ is decidable. 
\begin{proof}
	We restrict our attention to derivations in $ \mathsf{SC}^{?}$ in which no sequent appears twice in a branch, following the usual reduction to \textit{concise} proofs. When confronted with an end-sequent $ \Gamma\sststile{\mathbf{S}}{}\Delta $, we must first determine whether it is defeated. If it is, then there is no proof. If it is undefeated, then we proceed in the manner of the standard \textit{proof search algorithm} for sequent calculi, i.e. we consider all the possible inference rules that could have $ \Gamma\sststile{\mathbf{S}}{}\Delta $ as a conclusion and construct a number of trees, one for each distinct possibility, writing down the premises of such rules above it. Matters are slightly complicated by the presence of the $ \mathsf{DE} $ rule. However, since the defeater sets of any sequent are finite, we know that there is an upper bound to the number of derivations for any end-sequent containing a nonempty defeater set. We then repeat this process for the premises of each successive rule, at each point checking to see whether the premises are defeated. If a premise is defeated in a tree, then, following Definition \ref{def:proof}, it is not a possible proof, and thus we turn to the remaining trees. We proceed in this fashion until we have trees with no defeated sequents whose leaves are axioms.
\end{proof}
\end{theorem}

In order to reduce the length of proofs---a desideratum for any implementation of the system---we can add a \textit{cut} rule to the calculus, i.e: 

$$\begin{prooftree}[template = \small$\inserttext$]
\Hypo{\Gamma\sststile{\mathbf{S}}{}F,\Delta}
\Hypo{\Gamma', F\sststile{\mathbf{T}}{}\Delta'}
\Infer2[$cut$]{ \Gamma',\Gamma\sststile{\mathbf{S}\,\cup\,\mathbf{T}}{}\Delta,\Delta'}
\end{prooftree}$$

\noindent$ \mathsf{SC^?}$ is sound and complete for CPL and, as noted above, the relation of e$_{sr}$-implication is transitive, so this rule is admissible when the cut formula is either a d-wff or an e-wff.

\section{Example}\label{sec:ex}

Assume that an agent, $ \alpha $ is assigned the task of determining which of the following formulas is true: $ p, q $. Assume further, that $ \alpha $ has learned that $ \neg s \vee p $ and $s\vee q $ but has yet to obtain an answer to its principal question. Additionally, $ \alpha $'s knowledge base includes the information about defeaters. Writing `$ \rightsquigarrow $' for `defeats,' the information is as follows: $ \{r\}\rightsquigarrow s $, $ \{t\}\rightsquigarrow p $ and $ \{u, v\} \rightsquigarrow q $. Now, what question(s) should $ \alpha $ ask? There are various algorithms or heuristics the agent might operate with that determine or suggest ways of using $\mathsf{SC}^{?}$ to arrive at strategies for answering principal questions via sub-questions. Formulating such rules is an interesting exercise in its own right, but suppose that $ \alpha $ is provided with the following. 

\textbf{Erotetic Strategy:} For a question, $ Q $, and a set of facts, $ X $, determine whether you are licensed to inquire into $ ? A $ for any formula, $ A $, that occurs as a subformula in member of $ X $ but not in a member of $ dQ $, i.e. find a question, distinct from $ Q $, that is e$_{rs}$-implied by $ Q\cup X $.

\textbf{Strategy Implementation in $ \mathsf{SC}^? $:}  Try to prove
$ X, Q \sststile{\mathbf{S}\cup\,[d\mathbb{Q}]}{}?A $ such that $ A \in \mathsf{Subform}(X)\backslash\mathsf{Subform}(dQ) $.


The following proof realizes this strategy.
$$\begin{prooftree}[template = \scriptsize$\inserttext$, separation = .5em]
\Hypo{}
\Infer1[${\scriptstyle ax_3}$]{ \quad\sststile{\{\{r\}\}}{} s, \neg s}
\Infer1[$ {\scriptstyle \mathsf{LW}}$]{ ?\{p, q\}\sststile{\{\{r\}\}}{} s, \neg s }
\Hypo{}
\Infer1[$ {\scriptstyle ax_4}$]{ s, \neg s\sststile{\emptyset}{} \quad }
\Hypo{}
\Infer1[$ {\scriptstyle ax_1}$]{ p\sststile{\{\{t\}\}}{} p }
\Infer2[$ {\scriptstyle {\vee}{\vdash}}$]{ \neg s \vee p, s \sststile{\{\{t\}\}}{} p }
\Hypo{}
\Infer1[$ {\scriptstyle ax_4}$]{ s, \neg s\sststile{\emptyset}{} \quad }
\Hypo{}
\Infer1[$ {\scriptstyle ax_1}$]{ q\sststile{\{\{u, v\}\}}{} q }
\Infer2[$ {\scriptstyle {\vee}{\vdash}}$]{ s \vee q, \neg s \sststile{\{\{u, v\}\}}{} q }
\Infer3[$\vdash?_{\scriptscriptstyle 2}$]{?\{p, q\}, \neg s \vee p, s\vee q\sststile{\{\{r\},\{t\},\{u, v\}, \{p\}, \{q\}\}}{}\: ?\{s, \neg s\}}
\end{prooftree}$$

With the derivation of the end-sequent above, $ \alpha $ now has an erotetic inference that licenses it to inquire into $?\{s, \neg s\}$ as a means of answering $ ?\{p, q\} $. There are a few ways to conceive of $ \alpha $'s use of this inference. Perhaps the simplest is that once its derivation is obtained, $ \alpha  $ transitions from its inference mode to data-collection, and back again, adding each new piece of information to the sequent's antecedent via $ \mathsf{LW} $ until the sequent is defeated. Such defeat then represents the fact that either $ \alpha $'s initial question has been answered (i.e. if the new antecedent implies $ p $ or $q$) or that exceptions to the inference's warrant have been found (i.e. if the new antecedent implies $\{r\},\{t\},$ or $\{u, v\}$). 

\section{Conclusion}\label{sec:concl}

The calculus $ \mathsf{SC}^? $ satisfies our desiderata of an informally adequate and formally well-behaved calculus for defeasible erotetic inferences. The calculus' decidability makes it suitable for automated reasoning systems, such as ATPs discussed in the introduction. We think the system lays the ground for future investigations into the nature of erotetic defeasibility and the implementation of zetetically rational agents.

There are several ways to augment or otherwise alter $\mathsf{SC}^{?}$ that might capture other types of erotetic inference or features of erotetic defeat. For instance, the defeater set mechanism is rather coarse-grained insofar as it assimilates information that undermines propositional inferences with that which undermines erotetic inferences, i.e. answers. A simple way to distinguish and track changes in these two data sets would be to adorn the turnstile with two defeater sets, one that retains the sets of initial sequents, and one that only encodes additions made via one of the e-wff rules. Defeater sets would still be assembled in the binary rules via set-union.

Another way in which $\mathsf{SC}^{?}$ might be fruitfully developed would be to extend it in a such a way as to capture the general relation of erotetic implication, where implied questions need only entail proper subsets of answers to the implying question. To do so, we might let $ \Theta_i, \Theta_j $ stand for nonempty proper subsets of $ [ A_{\mid n}] $, i.e., $ \Theta_i, \Theta_j \in \mathcal{P}([ A_{\mid n}])/\{\emptyset, [ A_{\mid n}]\} $. By replacing $ A_i, A_j $ in $ {\vdash}{?_2} $ and $ {?}{\vdash_2} $ with $ \Theta_i, \Theta_j$, we obtain a rule that yields sequents that are sound and complete with respect to general erotetic implication. 

It would be quite interesting to see how $\mathsf{SC}^{?}$ compares to IEL's \textit{erotetic search scenarios}, which model the behavior of interrogators who answer a principle question by seeking out more easily or economically obtained answers to subordinate questions. Since strategies for resolving these scenarios are based on e-implicative relations, there are provable sequents in $\mathsf{SC}^{?}$ corresponding to each successful erotetic move. As algorithms for generating these scenarios have already been implemented \citep{Chlebowski2017,Bolotov2006,Leszczyska-Jasion2013,Lupkowski2015}, there is reason to think that implementing $\mathsf{SC}^{?}$ is both feasible and complementary of the systems for erotetic search scenarios. This would be one of the many ways to advance the study and design of zetetic agents.

\renewcommand{\bibfont}{\small}
\setlength{\bibsep}{6pt}

%
%
%
%
%
%

\newpage
\begin{figure}[!ht]  
\caption{Rules for $\mathsf{SC}^\mathbf{S} $}\label{fig:SCd}
\hspace{2mm}

\textbf{Axioms}
\vspace{.2cm}

\hspace{2mm}
$ 	\begin{prooftree}[template = \small$\inserttext$]
\Hypo{}
\Infer1[$ ax_1$]{ p \sststile{\mathcal{S}_p }{} p}
\end{prooftree}
\hspace{.7cm}
\begin{prooftree}[template = \small$\inserttext$]
\Hypo{}
\Infer1[$ ax_2$]{ \neg p \sststile{\mathcal{S}_p }{} \neg p}
\end{prooftree}
\hspace{.7cm}
\begin{prooftree}[template = \small$\inserttext$]
\Hypo{}
\Infer1[$ ax_3$]{\quad\sststile{\mathcal{S}_p }{} p, \neg p}
\end{prooftree} 
\hspace{.7cm}
\begin{prooftree}[template = \small$\inserttext$]
\Hypo{}
\Infer1[$ ax_4$]{  p, \neg p\sststile{\emptyset}{}\quad}
\end{prooftree} $

\vspace{.5cm}
\textbf{Structural Rules}
\vspace{.3cm}

\hspace{2mm}
$\begin{prooftree}[template = \small$\inserttext$]
\Hypo{\Gamma\sststile{\mathbf{S}}{}\Delta}
\Infer1[ $ \mathsf{LW} $ ]{ \Gamma, A\sststile{\mathbf{S}}{}\Delta}
\end{prooftree}$
\hspace{1.5cm}
$\begin{prooftree}[template = \small$\inserttext$]
\Hypo{\Gamma\sststile{\mathbf{S}}{}\Delta}
\Infer1[ $ \mathsf{RW} $ ]{ \Gamma\sststile{\mathbf{S}}{}A, \Delta}
\end{prooftree}$
\hspace{1.5cm}
$\begin{prooftree}[template = \small$\inserttext$]
\Hypo{\Gamma\sststile{\mathbf{S}}{}\Delta}
\Infer1[ $ \mathsf{DE} $ ]{ \Gamma\sststile{\mathbf{S}\,\cup\,\mathbf{T}}{}\Delta}
\end{prooftree}
$

\vspace{.5cm}
\textbf{Logical Rules}
\vspace{.3cm}

\hspace{2mm}
$\begin{prooftree}[template = \small$\inserttext$]
\Hypo{\Gamma,A,B\sststile{\mathbf{S}}{}\Delta}
\Infer1[${\wedge}{\vdash}$]{ \Gamma,A\wedge B\sststile{\mathbf{S}}{}\Delta}
\end{prooftree}$
\hspace{3.5cm}
$\begin{prooftree}[template = \small$\inserttext$]
\Hypo{\Gamma\sststile{\mathbf{S}}{}A,\Delta}
\Hypo{\Gamma'\sststile{\mathbf{T}}{}B,\Delta'}
\Infer2[${\vdash}{\wedge}$]{ \Gamma',\Gamma\sststile{\mathbf{S}\,\cup\, \mathbf{T}}{}A\wedge B,\Delta,\Delta'}
\end{prooftree} $

\vspace{.5cm}
\hspace{2mm}
$\begin{prooftree}[template = \small$\inserttext$]
\Hypo{\Gamma,A\sststile{\mathbf{S}}{}\Delta}
\Hypo{\Gamma',B\sststile{\mathbf{T}}{}\Delta'}
\Infer2[$ {\vee}{\vdash}$]{ \Gamma',\Gamma, A\vee B\sststile{\mathbf{S}\,\cup\,\mathbf{T}}{}\Delta,\Delta'}
\end{prooftree}$
\hspace{3.5cm}
$\begin{prooftree}[template = \small$\inserttext$]
\Hypo{\Gamma\sststile{\mathbf{S}}{}A,B,\Delta}
\Infer1[${\vdash}{\vee}$]{ \Gamma\sststile{\mathbf{S}}{}A\vee B,\Delta}
\end{prooftree} $

\vspace{.5cm}

\hspace{2mm}
$\begin{prooftree}[template = \small$\inserttext$]
\Hypo{\Gamma, A\sststile{\mathbf{S}}{}\Delta}
\Infer1[${\neg}{\neg}{\vdash}$]{ \Gamma,\neg\neg A\sststile{\mathbf{S}}{}\Delta}
\end{prooftree}$
\hspace{5cm}
$\begin{prooftree}[template = \small$\inserttext$]
\Hypo{\Gamma\sststile{\mathbf{S}}{}A,\Delta}
\Infer1[${\vdash}{\neg}{\neg}$]{\Gamma\sststile{\mathbf{S}}{}\neg \neg A, \Delta}
\end{prooftree} $
\vspace{.5cm}

\hspace{2mm}
$\begin{prooftree}[template = \small$\inserttext$]
\Hypo{\Gamma, \neg A\sststile{\mathbf{S}}{}\Delta}
\Hypo{\Gamma', \neg B\sststile{\mathbf{T}}{}\Delta'}
\Infer2[${\neg}{\wedge}{\vdash}$]{ \Gamma',\Gamma, \neg (A\wedge B)\sststile{\mathbf{S}\,\cup\, \mathbf{T}}{}\Delta,\Delta'}
\end{prooftree}$ 
\hspace{2.2cm}
$\begin{prooftree}[template = \small$\inserttext$]
\Hypo{\Gamma\sststile{\mathbf{S}}{}\neg A,\neg B,\Delta}
\Infer1[${\vdash}{\neg}{\wedge}$]{ \Gamma\sststile{\mathbf{S}}{}\neg(A \wedge B),\Delta}
\end{prooftree}
$

\vspace{.5cm}
\hspace{2mm}
$\begin{prooftree}[template = \small$\inserttext$]
\Hypo{\Gamma,\neg A,\neg B\sststile{\mathbf{S}}{}\Delta}
\Infer1[${\neg}{\vee}{\vdash}$]{ \Gamma,\neg(A \vee B)\sststile{\mathbf{S}}{}\Delta}
\end{prooftree}$
\hspace{2.2cm}
$\begin{prooftree}[template = \small$\inserttext$]
\Hypo{\Gamma\sststile{\mathbf{S}}{}\neg A,\Delta}
\Hypo{\Gamma'\sststile{\mathbf{T}}{}\neg B,\Delta'}
\Infer2[${\vdash}{\neg}{\vee}$]{ \Gamma',\Gamma\sststile{\mathbf{S}\,\cup\, \mathbf{T}}{}\neg(A\vee B),\Delta,\Delta'}
\end{prooftree} $

\vspace{.5cm}

\end{figure}

\begin{landscape}
\begin{figure}[!h]  
	\caption{Rules for $\mathsf{SC}^{?}$}\label{fig:SC?}
	\hspace{2mm}
	
	\textbf{Logical Axioms} 
	\vspace{.2cm}\\
	\hspace*{2mm}Same as $\mathsf{SC}^\mathbf{S} $
	
	\hspace{2mm}
	
	\textbf{Structural Rules}
	\vspace{.3cm}
	
	\hspace{2mm}
	$ 	\begin{prooftree}
	\Hypo{\Gamma\sststile{\mathbf{S}}{}\Delta}
	\Infer1[ $ \mathsf{LW} $ ]{ \Gamma, F\sststile{\mathbf{S}}{}\Delta}
	\end{prooftree}
	\hspace{2cm}
	\begin{prooftree}[template = \small$\inserttext$]
	\Hypo{\Gamma\sststile{\mathbf{S}}{}\Delta}
	\Infer1[ $ \mathsf{RW} $ ]{ \Gamma\sststile{\mathbf{S}}{}F, \Delta}
	\end{prooftree}
	\hspace{2cm}
	\begin{prooftree}[template = \small$\inserttext$]
	\Hypo{\Gamma\sststile{\mathbf{S}}{}\Delta}
	\Infer1[ $ \mathsf{DE} $ ]{ \Gamma\sststile{\mathbf{S}\,\cup\,\mathbf{T}}{}\Delta}
	\end{prooftree}
	$
	
	\vspace{.5cm}
	\textbf{Logical Rules for D-wffs}
	\vspace{.3cm}\\
	\hspace*{2mm}
	Same as Logical Rules in $\mathsf{SC}^\mathbf{S}$
	
	\vspace{.3cm}
	
	\textbf{Logical Rules for E-wffs}
	
	\vspace{.3cm}
	
	\begin{center}
		\hspace{2mm}
		\begin{prooftree}
			\Hypo{ X\sststile{\mathbf{S} }{} A_1,\ldots,A_n, \Delta}
			\Infer1[${\vdash}{?_{\scriptscriptstyle 1}}^\dag$]{X\sststile{\mathbf{S}\,\cup\,\{\{A_1\},\ldots,\{A_n\}\}}{}
				\: ?\{A_1,\ldots,A_n\}, \Delta}
		\end{prooftree}	\hspace{1cm}
		\begin{prooftree}
			\Hypo{\Gamma_1, A_1 \sststile{\mathbf{S}_1}{} X_1}
			\Hypo{\hspace{-5.5mm}\ldots\Gamma_n, A_n \sststile{\mathbf{S}_n }{} X_n}
			\Infer2[${?}{\vdash_{\scriptstyle 1}}^\dag$]{ \Gamma_{\mid n}, ?\{A_1,\ldots,A_n\} \sststile{\mathbf{S}_{\mid n}\,\cup\,\{\{A_1\},\ldots,\{A_n\}\}}{}X_{\mid n}}
		\end{prooftree}
		
		\vspace{.4cm}
		
		\begin{prooftree}
			\Hypo{\Gamma_1, ?[A_{\mid n}]\sststile{\mathbf{S}}{}[B_{\mid m}], \Delta_1}
			
			\Hypo{\Gamma_2, B_{1}\sststile{\mathbf{T}_{1} }{} A_i, \Delta_2}
			\Hypo{\hspace{-5mm}\ldots\Gamma_{m+1}, B_{m}\sststile{\mathbf{T}_{m} }{} A_{j}, \Delta_{m+1}}
			\Infer3[${\vdash}{?_{\scriptstyle 2}}\,^\ddag$]{ \Gamma_{\mid m+1} , ?[A_{\mid n}] \sststile{\mathbf{S}\,\cup\,\mathbf{T}_{\mid m}\cup\,\{\{A_1\},\ldots,\{A_n\}\}}{}  \,?[B_{\mid m}], \Delta_{\mid m+1}}
		\end{prooftree}
		
		\vspace{.4cm}
		
		\begin{prooftree}
			\Hypo{\Gamma_{1}, A_{1}\!\sststile{\mathbf{S}_{1} }{}\,?[B_{\mid m}], \Delta_1 \:\ldots}
			\Hypo{\hspace{-4mm}\Gamma_{n}, A_{n}\!\sststile{\mathbf{S}_{n} }{}\,?[B_{\mid m}], \Delta_n }
			\Hypo{\hspace{4mm}\Gamma_{n+1}, B_{1}\!\sststile{\mathbf{T}_{1} }{} A_i , \Delta_{n+1} \:\ldots}
			\Hypo{\hspace{-4mm}\Gamma_{n+m}, B_{m}\!\sststile{\mathbf{T}_{m} }{} A_j, \Delta_{n+m}}
			\Infer4[$\!\!{?}{\vdash_{\scriptstyle 2}}\,^{\!\!\ddag}$]{\Gamma_{\mid n+m}, ?[A_{\mid n}] \sststile{\mathbf{S}_{\mid n}\cup\,\mathbf{T}_{\mid m}\cup\,\{\{A_1\},\ldots,\{A_n\}\}}{}\, ?[B_{\mid m}], \Delta_{\mid n+m}}
		\end{prooftree}
		
	\end{center}
	
	\vspace*{3mm}
	$ \dag $ Provided $ n>1 $ and $ A_1,\ldots,A_n $ are non-equiform.
	
	$ \ddag $ Provided $\{A_{i},\ldots,A_{j}\} \subseteq \{ A_1,\ldots,A_n \}$; $ m,n>1$ ;  $ A_1,\ldots,A_n $ are pairwise non-equiform; and  $ B_1,\ldots,B_n $ are pairwise non-equiform.
	
\end{figure}
\end{landscape}


\noindent{\bseries Acknowledgments.} I would like to thank an anonymous referee for suggestions regarding earlier versions of this paper.

\AuthorAdressEmail{Jared Millson}{Department of Philosophy \& Religious Studies\\ California State University---Bakersfield\\
Bakersfield, CA USA}{jmillson@csub.edu}



\label{k}
\end{document}